\RequirePackage{fix-cm}
\documentclass[smallextended]{svjour3}       % onecolumn (second format)

\smartqed  % flush right qed marks, e.g. at end of proof
\usepackage{graphicx}
\usepackage{hyperref}
\usepackage{mathptmx}      % use Times fonts if available on your TeX system
\usepackage{tabularx}
\usepackage{booktabs}
\usepackage{xcolor}
\makeatletter
\def\cl@chapter{\@elt {theorem}}
\makeatother

\usepackage[capitalize]{cleveref}
\usepackage{subcaption}
\newcommand{\paretoFig}[2]{
\begin{figure}[tbh]
		\vspace{-.5em}
	\centering
	\includegraphics[width=.865\linewidth]{figures/#1-PLOTS}
	\vspace{-.5em}
	\caption{Approximated Pareto front on the \textit{#2} dataset.}
		\vspace{-1em}
	\label{fig:#1-plots}
\end{figure}
}

\newcommand{\compareFig}[2]{
	\begin{figure}[tbh]
		\vspace{-.5em}
		\centering
		\includegraphics[width=.865\linewidth]{figures/#1-compare-PLOTS}
		\vspace{-.5em}
		\caption{Classification performance compared to benchmark methods on the \textit{#2} dataset.}
		\vspace{-1em}
		\label{fig:#1-compare-plots}
	\end{figure}
}

\newcommand{\revisionOne}[1]{{#1}}
\newcommand{\revisionOnePar}[1]{{#1}}
\journalname{Genetic Programming and Evolvable Machines}
\begin{document}
\title{Multi-Objective Genetic Programming for Manifold Learning: Balancing Quality and Dimensionality}
%\title{Multi-objective Manifold Learning using \\Genetic Programming%\thanks{Grants or other notes

\titlerunning{Multi-Objective Genetic Programming for Manifold Learning}        % if too long for running head

\author{Andrew Lensen         \and
        Mengjie Zhang \and
        Bing Xue %\and
        %Mengjie Zhang
}


\institute{Andrew Lensen, Mengjie Zhang, Bing Xue \at
              School of Engineering and Computer Science, \\
              Victoria University of Wellington, PO Box 600, Wellington 6140, New Zealand\\
              %Fax: +123-45-678910 \\
              Email: \href{mailto:andrew.lensen@ecs.vuw.ac.nz}{\{andrew.lensen,bing.xue,mengjie.zhang\}@ecs.vuw.ac.nz}
}

\date{Received: date / Accepted: date}

\maketitle

\begin{abstract}
Manifold learning techniques have become increasingly valuable as data continues to grow in size. By discovering a lower-dimensional representation (embedding) of the structure of a dataset, manifold learning algorithms can substantially reduce the dimensionality of a dataset while preserving as much information as possible. However, state-of-the-art manifold learning algorithms are opaque in how they perform this transformation. Understanding the way in which the embedding relates to the original high-dimensional space is critical in exploratory data analysis. We previously proposed a Genetic Programming method that performed manifold learning by evolving mappings that are transparent and interpretable. This method required the dimensionality of the embedding to be known \textit{a priori}, which makes it hard to use when little is known about a dataset. In this paper, we substantially extend our previous work, by introducing a multi-objective approach that automatically balances the competing objectives of manifold quality and dimensionality. Our proposed approach is competitive with a range of baseline and state-of-the-art manifold learning methods, while also providing a range (front) of solutions that give different trade-offs between quality and dimensionality. Furthermore, the learned models are shown to often be simple and efficient, utilising only a small number of features in an interpretable manner. 

\end{abstract}

\section{Introduction}
\label{intro}
Manifold learning (MaL) has become a subject of significant research interest in recent years, with 14,100 occurrences of the phrase in research papers since 2015 alone, out of the 29,500 total all-time mentions\footnote{Data gathered using \url{https://scholar.google.com/scholar?q=\%22manifold+learning\%22} \revisionOne{on 15th October, 2019}.}. MaL, also called nonlinear dimensionality reduction, is an unsupervised learning problem based on the observation that most high-dimensional data is only artificially high --- that is, most data can be represented in a much lower-dimensional space by discovering the intrinsic structure within a dataset, called an (embedded) \textit{manifold} \cite{cayton2005algorithms}. MaL algorithms use this assumption to attempt to discover the best low-dimensional embedding that can be found for a dataset, such that as much high-dimensional structure as possible is maintained in the lower-dimensional space.

A common categorisation of MaL algorithms is based on whether they are ``embedding'' or ``mapping'' methods \cite{lee2007nonlinear}. Embedding methods directly optimise the low-dimensional manifold by treating the task essentially as a numerical optimisation problem, typically producing only the final embedding as a result. Mapping methods, in contrast, provide a mapping or \textit{model} that converts the original feature space into the low-dimensional manifold.  These mapping methods have two particularly desirable properties: they can be re-used on future instances without having to re-optimise; and they have the potential to be interpretable in terms of the original high-dimensional features, which is of particular importance in many exploratory data mining tasks \revisionOne{such as medical diagnosis, automatic credit score calculation, and trustworthiness of climate change models} \cite{murdoch2019interpretable}. Mapping methods also have the potential to be more efficient/higher-performing than embedding ones: on datasets with a very large number of instances, it becomes difficult to optimise each instance's embedded values directly, and so learning a mapping for all instances may be more viable. Despite this, there have been very few mapping methods proposed in the literature, with most focus in recent years on gradient-based embedding methods \cite{maatenTSNE,2018arXivUMAP}.

Evolutionary computation (EC) methods have seen very little use for manifold learning. This is despite MaL displaying all the hallmarks of a task well-suited to EC: it is NP-hard, easy to model as an optimisation problem, and has local optima in the form of the most dominant patterns in the manifold. Recently, we proposed GP-MaL \cite{lensen2019can}, the first approach to using genetic programming (GP) to automatically evolve mappings in manifold learning. GP-MaL showed clear potential when compared to other MaL methods. However, GP-MaL required that the number of dimensions in the embedding (the number of trees in a GP individual) be known in advance. For true exploratory data analysis, this information is unlikely to be available. In this paper, we aim to develop GP-MaL further by proposing a multi-objective (MO) version (GP-MaL-MO) that can automatically find a good trade-off between the dimensionality of the embedded manifold and the amount of structure that is maintained within it. More specifically, this paper will:

\begin{itemize}
	\item Introduce a revised GP multi-tree representation that allows for a variable number of trees in an individual, as well as new crossover and mutation operators that allow for the addition/removal of trees;
	\item Formulate GP-based MaL as a multi-objective problem with two objectives representing manifold quality and manifold dimensionality;
	\item Evaluate the trade-off (Pareto front) between these tasks on a variety of commonly-used datasets;
	\item Compare GP-MaL-MO to existing MaL methods across a range of different embedded manifold sizes; and
%	\item Further investigate the potential of GP-MaL-MO for producing interpretable, representative manifolds.
	\item \revisionOne{Investigate if GP-MaL-MO has the potential to evolve interpretable manifolds.}
\end{itemize}

%\begin{itemize}
%	\item Basic intro to manifold learning
%	\item Mapping vs embedding methods
%	\item Motivation for using GP
	%\item Highlight EuroGP paper using GP. But that the number of trees needed to be set in advance which wasn't always obvious. More trees $\propto$ more structure of data retained (lower cost), but clear trade-off vs number of dimensions in embedding. So multi-objective problem: we can find a Pareto front to optimise.
	%\item This paper does so, using EMO GP.
%\end{itemize}

\section{Background}
In this section, we introduce the key fields of unsupervised learning, dimensionality reduction, and manifold learning. We then discuss a selection of related work in this domain.
\subsection{Unsupervised Learning}
Unsupervised learning \cite{hastie2009unsupervised} is a category of machine learning in which there are no pre-existing \textit{labels} (or ``guidance'') available for datasets. In contrast to supervised learning, which attempts to learn to re-create the provided labels, in unsupervised learning algorithms must discover inherent structure or useful intrinsic patterns within a dataset.

The most common and prototypical example of an unsupervised learning problem is cluster analysis \cite{aggarwal2014data}, where data is grouped into a number of clusters that contain instances that are closely related by some aspect of their feature distributions. Other common tasks include anomaly detection \cite{chandola2009anomaly}, blind signal separation \cite{comon2010handbook}, and --- core to this work --- nonlinear dimensionality reduction \cite{lee2007nonlinear}. 
%Without labels; clustering is common, manifold learning is more recent and is different.
\subsection{Dimensionality Reduction}
Dimensionality Reduction (DR) is, simply put, the task of reducing a high-dimensional space (or dataset) to a smaller one \cite{liu2012feature}. DR techniques are most commonly used when datasets are too large to be used directly with existing machine learning techniques, or when we wish to \textit{understand} the structure of a dataset, which is infeasible at even relatively small dimensionality. 

The two most common DR approaches are feature selection and feature construction (FC) (or extraction) \cite{liu1998feature}. FS is the more intuitive of the two: FS methods seek to find a subset of the original feature set that maintains properties of the data sufficiently well. For example, in a classification task, FS may be performed by finding the smallest subset of features possible that does not result in a meaningful decrease in classification accuracy \cite{liu2005toward}. FC methods take a different approach: they create (\textit{construct}) new meta-features, which are a combination of a number of features in the original feature set. FC methods may use a range of different operators (functions) to combine a subset of features into a new, higher-level feature. By doing this, they have the potential to remove several features at once, replacing them with a single (or multiple) constructed features that maintain useful information from all the features used in their construction. FS methods tend to be simpler to design and more computationally efficient to run, but are fundamentally limited in how much they can reduce dimensionality. FC methods, in contrast, can theoretically reduce dimensionality more dramatically --- but it is difficult to find good combinations of features given the huge number of possible combinations.

FS and FC have seen wide application in supervised learning tasks \cite{tang2014feature,sondhi2009fc}, including EC-based FS \cite{xue2015survey} and FC \cite{neshatian2012filter,tran2016genetic}. FS and FC have also been used in unsupervised learning tasks \cite{dy2004feature}, but little research has considered EC-based approaches for these tasks \cite{emlSurvey2019}. In particular, only a few papers have used EC-based FC for unsupervised learning: mostly in clustering tasks \cite{boric2007genetic,coelho2011multi}, but also for the creation of benchmark datasets \cite{shand2019evolving,lensen2018automatically}. %Until our recent work \cite{lensen2019can}, there was no work that directly used EC methods to perform manifold learning.

\subsection{Manifold Learning}
Manifold learning is the task of discovering the inherent structure within a high-dimensional dataset, so that that it can be represented by a much lower-dimensional space \cite{bengio2013representation}. MaL is an unsupervised learning problem, as only the relationships between instances are considered in searching for this structure. ``Embedding'' MaL methods can be regarded as a type of optimisation problem, where the feature values of the low-dimensional space (embedding) are optimised according to some cost function based on the high-dimensional space. The ``mapping'' MaL methods, however, naturally represent a FC problem: given a number of features (the high-dimensional space), the method must search for the best way to combine the these features into a low-dimensional space, using \revisionOne{a} cost function to measure the quality of a mapping. 

It is this observation that is central to our use of GP for manifold learning. A program evolved by GP takes some number of inputs, and produces an output(s). This is, of course, a function from one set of (input) features to one or more output features. In \revisionOne{GP-MaL} \cite{lensen2019can}, we showed this observation can be used in practice, by learning a number of GP trees to produce a manifold.

A wide variety of MaL methods have been proposed, and we refer interested readers to the survey conducted by Bengio et al.\ \cite{bengio2013representation} for an in-depth overview of the field. 
%Specialised type of FC in some sense, briefly mention common approaches/algorithms.
\subsection{Related Work}
GP-MaL \cite{lensen2019can}, was the first approach proposed that used an EC technique as a mapping MaL method. We showed that GP-MaL was clearly competitive with other MaL methods, including non-mapping methods. In particular, the ability of GP-MaL to create relatively simple and efficient mappings, which had the potential to be interpreted, was a key advantage over other MaL methods.

EC has been integrated with MaL tasks in other ways, such as aiding in the evolution of autoencoders, using Genetic Algorithms \cite{sun2018evolving}, GP \cite{mcdermott2019why,rodriguez2018structurally}, and Particle Swarm Optimisation \cite{hsu2017unsupervised}. The use of EC for aiding in these tasks is clearly beneficial, as neural networks (which the autoencoder is a type of) have long struggled from needing intensive human design and having more complex structures than necessary \cite{floreano2008neuroevolution}. In this paper, we are primarily interested in using GP to \textit{directly} perform manifold learning --- rather than via another algorithm --- in the hope that it will better take advantage of the FC abilities of GP. 

Asides from the evolution of autoencoders, there are not any other directly comparable areas of research. Some other tasks are tangentially related, such as visualisation. Visualisation can be considered as a two-dimensional manifold learning problem. For example, GP has been used to visualise the quality of solutions in job shop scheduling \cite{nguyen2018visualising}, and as one criteria in a multi-objective GP classification problem \cite{icke2011multi}. \revisionOne{There has also been recent research on using non-EC manifold learning to visualise EC methods themselves \cite{lorenzo2019an,michalak2019low}.} 
%\begin{itemize}
%	\item Dimensionality reduction $\to$ FC $\to$ GP.
%	\item FC $\equiv$ Manifold learning.
%	\item Related work: GP-MaL (EuroGP), as well as other autoencoder etc papers.
%\end{itemize}
\section{Genetic Programming for Manifold Learning using a Multi-objective Approach (GP-Mal-MO)}
%Our previous work, GP-MaL \cite{lensen2019can}, proposed the first GP-based approach to performing manifold learning. 
In this paper, we propose a multi-objective extension to \revisionOne{GP-MaL \cite{lensen2019can}}, named GP-MaL-MO. While the core of the algorithm remains the same as in GP-MaL, a few changes were made to further improve results. As such, we describe the design of GP-MaL-MO fully in this section.
 
\subsection{GP Representation}
We utilise the same fundamental GP representation as in \revisionOne{GP-MaL}: a multi-tree GP, where each individual \revisionOne{$I$ contains a list of $t = |I|$ trees, $I = [T_1,T_2,...,T_t]$, each of which represents a single dimension in the low-dimensional manifold}. In other words, each tree is a single constructed feature, which represents some part of the high-dimensional structure of the dataset. Unlike in GP-MaL, GP-MaL-MO does not fix $t$ in advance for all individuals in the population. Instead, it randomly initialises the number of trees in each individual, and then allows individuals to gain/lose trees throughout the evolutionary process via a mutation operator.

The terminal set consists only of the $d$ real-valued input features, each uniformly scaled to the range [0,1] to reduce bias. Unlike in GP-MaL, we do not include ephemeral random constants as their removal did not reduce performance. \revisionOne{Theoretically, there should be no need for ERCs in manifold learning, as the high-dimensional structure is defined entirely by the combination of the original features themselves. The removal of ERCs also helps to reduce the search space of possible trees, and may aid tree interpretability as sub-trees will now have equal weighting.} The function set, shown in \cref{functionSet}, is also very similar to the one in GP-MaL. Based on \revisionOne{further testing}, the two-input addition function was removed, and the division and subtraction operators added to the function set. As in GP-MaL, we use the arithmetic functions to allow simple combinations of sub-trees; the non-linear functions to aid in producing more flexible, non-linear manifolds; and the conditional functions to allow representing varied structure in across parts of the data. The division operator is protected: it returns 1 if the denominator is 0.

\begin{table}[tb]
	\centering
	%\vspace{-1.25em}
	\caption{Summary of the function set used in GP-MaL-MO.}
	\label{functionSet}
	%	\setlength\tabcolsep{2mm}
	%\resizebox{\textwidth}{!}{%
	\begin{tabularx}{\textwidth}{@{}lXllllXccXlll@{}}
		\toprule
		Category  & & \multicolumn{4}{c}{Arithmetic} && \multicolumn{2}{c}{Non-Linear} && \multicolumn{3}{c}{Conditional} \\ \cmidrule{3-6}\cmidrule{8-9}\cmidrule{11-13} %\midrule
		Function &  & $5+$    & $-$ & $\times$   & $\div$ &    & Sigmoid    & ReLU      &   & Max       & Min     & If      \\ 
		
		No.\ of Inputs & & 5      & 2     & 2      & 2    &   & 1               & 1        &    & 2         & 2         & 3       \\
		No.\ of Outputs && 1      & 1      & 1     & 1     &  & 1               & 1        &    & 1         & 1         & 1       \\ \bottomrule
	\end{tabularx}%
	%	}
%	\vspace{-1.5em}
\end{table}

\subsection{Objective Functions}
GP-MaL-MO uses a multi-objective approach to approximate a Pareto front that gives trade-offs between the quality of a learned manifold and the number of dimensions required to achieve a given quality. Measuring the number of dimensions is straightforward: it is simply the number of trees ($t$) in an individual \revisionOne{$I$}. However, there are a plethora of approaches proposed for measuring how well structure has been preserved from a high- to a low-dimensional space \cite{bengio2013representation}. A common approach is to evaluate how well an instance's neighbourhood in the high-dimensional space matches that of the low-dimensional space. For example, if an instance $i$ has the three nearest neighbours \revisionOne{$i_1, i_2, i_3$} (in that order) in the high-dimensional space, then we would expect a good manifold learning algorithm to produce the same ordering in the low-dimensional space. 

\revisionOne{GP-MaL proposed a fitness function that measures the similarity between an instance's neighbour orderings in the high-dimensional space ($N$) and low-dimensional space ($N'$):}
\begin{equation}
Similarity(N, N') = \sum_{a \in N} Agreement(|Pos(a,N) - Pos(a,N')|)
%\vspace{-.5em}
\end{equation}
where $Pos(a,X)$ gives the index of neighbour $a$ in the ordering $X$, and the \textit{Agreement} function gives higher results for smaller deviations between the orderings. An \textit{Agreement} function based on a Gaussian weighting was used to punish larger deviations more harshly. The fitness was then calculated as the normalised similarity across all pairs of instances in the dataset.

GP-MaL-MO uses a similar approach, but instead of using a Gaussian weighting, we instead use Spearman's rank correlation coefficient \cite[p.~502]{dodge2008concise}. This eliminates the need to set the $\mu$ and $\theta$ hyper-parameters of the Gaussian weighting, while also being a more well-understood method of comparing rankings. As the orderings (rankings) of neighbours is ordinal, we compute the correlation coefficient between the two orderings as follows:
\begin{equation}
Correlation(N,N') = 1 - \frac{6 \sum_{i=1}^n d_i^2}{n(n^2-1)} 
\end{equation}
where $d_i = Pos(a,N) - Pos(a,N')$ \revisionOne{for instance $i$} and $n$ is the number of instances. Spearman's correlation has a range of $[-1,1]$, where a more negative value represents a stronger negative correlation between the two rankings, and a more positive value represents a stronger positive correlation. A value near zero represents a lack of correlation. When measuring how well the neighbourhood ordering is preserved, a positive correlation suggests neighbours have retained their ordering, whereas a negative correlation suggests that further-away neighbours have become closer, which is clearly a poor result. The number of trees should be minimised, and so it is easier to treat the neighbourhood preservation as a minimisation objective as well. Hence, we calculate the \textit{Cost} (lack of quality):
\begin{equation}
Cost(N,N') = \frac{\revisionOne{1}-Correlation(N,N')}{2}
\end{equation}
This gives a cost function in the range [0,1], where 0 represents a perfect preservation of an instance's neighbourhood ordering, and 1 a complete reversal. The cost of a given GP individual, $I$, is then the mean cost across all instances in the dataset ($X$):

\begin{equation}
\label{eqn:cost}
Cost(I) = \frac{1}{n} \sum_{x \in X} Cost(N_x,N_x')
\end{equation}

GP-MaL-MO hence aims to minimise both the two conflicting objectives of \textit{Cost} and the number of trees ($t$).

\subsection{Computational Complexity of Comparing Neighbourhoods}
The full neighbourhood of a given instance consists of every other instance in the dataset. Comparing the ordering of the full neighbourhood is quite expensive, as it requires a $O(n \log(n))$ comparison sort, which makes the complexity of evaluating the cost of a single GP individual $O(n^2 \log(n))$. In \revisionOne{GP-MaL} we introduced a sub-sampling technique, which computes only a partial ordering of a selection of neighbours. The same technique is used in GP-MaL-MO: we choose the first $k$ neighbours, followed by $k$ of the next $2k$ neighbours (spaced evenly), $k$ of the next $4k$ neighbours, and so on. This gives a sub-linear complexity of $O(\log(n) \log(\log(n)))$, which scales much more reasonably on large datasets. The reader is referred to our earlier work for more extensive discussion of this technique \cite{lensen2019can}.
 
\subsection{Evolutionary Operators}
GP-MaL-MO utilises specialised crossover and mutation operators that are designed to work effectively with varied numbers of trees.
\subsubsection{Crossover}  As $t$ is dynamic in GP-MaL-MO, we have altered the crossover operator slightly from GP-MaL. \revisionOne{As before, let us represent the trees of an individual $I$ as an indexed array $I = [T_1, T_2, ..., T_t]$ with length $t$}. 
In GP-Mal-MO, we restrict crossover to be between trees of the \textbf{same} index in each individual. This is enforced so as to encourage specialisation (i.e.\ heterogeneity) of trees in solutions. Consider two individuals representing two manifolds, each with the same number of dimensions ($t$). We would like to perform crossover such that trees representing similar structure of the data are picked to crossover together. If we randomly pick any tree from each individual, it is unlikely the best matching will be chosen for any reasonably high $t$. By restricting crossover to between the same indices, we naturally pressure trees of the same index to be related over the evolutionary process. When two individuals have different $t$ values, this restriction also encourages more sensible exchange of information between trees. An individual with a small $t$ will only be able to contain the ``global'' structure of the data, whereas one with a higher $t$ will be able to contain trees more specific to local structure in the data. This crossover technique encourages trees at a lower index to be ``global'' trees, and then trees at higher indices to be increasingly more ``local''. To further encourage specialisation, and to increase the learning rate of the EC process, we perform crossover between all valid pairs of trees in the two individuals. It follows that crossover is also restricted to occur only on the array indices common to both individuals. That is, given two individuals $A$ and $B$, crossover is performed in the range \revisionOne{up to and including the $i^{th}$ tree, where $i = \min\{A_t,B_t\}$, i.e.\ between the two (sub-)individuals $[A_1,A_2,...,A_i]$ and $[B_1,B_2,...,B_i]$.} 

%TODO: Reword: we do ALL pairs!!

\subsubsection{Mutation}  GP-Mal-MO utilises two types of mutation: one to alter the number of trees in an individual (``add/remove'' mutation), and one to alter trees within an individual (``standard'' mutation). Add/remove mutation is used to provide good solutions the chance to also become good solutions for similar ``sub-problems''. For example, a good solution containing seven trees may be able to lose one of these trees while still maintaining a relatively good (but worse) representation of the data. Conversely, it could gain an eighth tree, which may reduce the cost value sufficiently to warrant the added dimensionality. \revisionOne{More formally, an individual $I$ of length $t$ will be mutated (with equal probability) to a new length of either $t-1$ or $t+1$, with the constraint that $t-1 \geq 1$ and $t+1 \leq t_{max}$.} The ``standard'' GP mutation operator picks a tree $i$ at random in an individual \revisionOne{($i \in [1,t]$)}, picks a random sub-tree, and then generates a new sub-tree using the \textit{full} generation method.

\subsection{Choice of Multi-Objective Algorithm}
We use the Multiobjective Evolutionary Algorithm Based on Decomposition (MOEA/D) \cite{zhang2007moead} algorithm in this work for two main reasons. Firstly, we found it is particularly important to have strong diversity preservation as the EC process will naturally find it ``easier'' to generate non-dominated solutions with fewer trees (and higher cost) as it is difficult to find a large number of trees that can work well together, with a low cost value. MOEA/D is particularly well suited to diversity preservation due to its decomposition approach. 

The second reason is that MOEA/D also strongly encourages crossover to occur between solutions that represent similar trade-offs between the two objectives (i.e.\ similar sub-problems) through its use of a neighbourhood. This is again important to this work, as exchanging information between individuals with similar $t$ values is inherently more productive than those with dissimilar values of $t$, given that they are capable of preserving similar amounts of the dataset structure.

 %This ensures that trees which are appropriate for a manifold with a higher dimensionality are not shared with an individual representing a simpler manifold.
%If any pair of trees were able to crossover, 
\section{Experiment Design}
To evaluate the performance of GP-MaL-MO, we compare it to a selection of \revisionOne{five} existing manifold learning methods, using a variety of approaches. These include:

\begin{itemize}
\item Principal Component Analysis (PCA)\footnote{Technically, PCA does not perform manifold learning, as it does not perform non-linear dimensionality reduction, but it is still a useful baseline.} \cite{jolliffe2011pca}: computes a number of linearly uncorrelated components, such that each successive component represents the axis of most remaining variance;
\item MultiDimensional Scaling (MDS) \cite{kruskal1964mds}: attempts to maintain the high-dimensional distance between instances in the low-dimensional space;
\item Locally Linear Embedding (LLE) \cite{roweis2000lle}: models each instance as a linear combination of its high-dimensional nearest neighbours and attempts to maintain this combination in the low-dimensional space;
\item Uniform Manifold Approximation and Projection for Dimension Reduction (UMAP) \cite{2018arXivUMAP}: models the high-dimensional structure as a fuzzy topological structure, and then attempts to find a low-dimensional embedding that has the closest equivalent fuzzy topological structure. UMAP is often regarded as the state-of-the-art manifold learning technique, and a spiritual successor to the widely-known t-Distributed Stochastic Neighbour Embedding (t-SNE) method \cite{maatenTSNE} that we compared to in our initial work.
\item \revisionOne{A single hidden layer auto-encoder (AE) \cite{kramer1991ae}: auto-encoders are a specialised neural network used for (unsupervised) representation learning. In this work, we use a simple auto-encoder with a single hidden (embedding) layer that has the number of nodes equal to the number of desired created features. While more advanced AEs exist, they are generally outperformed by methods such as UMAP and are very difficult to interpret.}
\end{itemize}

\revisionOne{Each of the above methods are somewhat less expensive than GP-MaL-MO for the sizes of datasets tested: for example, MDS and LLE are approximately $O(n^2)$ (depending on implementation), whereas UMAP is $O(n^{1.14})$.  The major computational cost in GP-MaL-MO is the calculation of manifold quality at $O(\log(n) \log(\log(n)))$ for $n$ instances, as well as the initial full calculation of neighbours at $O(n^2)$. Given that the calculation of manifold quality must be performed for each individual in the population, for each generation, this gives a net complexity of $O(n^2 + G \times P \times \log(n) \log(\log(n)))$ for $G$ generations and a population size of $P$. However, it is necessary to run each of the baseline methods (except for PCA) once for each value of $t$, as opposed to GP-MaL-MO, which runs once.}

\revisionOne{We include the results of GP-MaL that were provided previously \cite{lensen2019can} (for some values of $t$). In contrast to GP-MaL-MO, which can produce solutions for a wide range of $t$ in a single run, GP-MaL requires $t$ to be chosen in advance. Running GP-MaL 30 times for every different $t$ produced by GP-MaL-MO would require an unreasonable amount of computational time, and so was not feasible to do.}

\subsection{Classification Accuracy as a Proxy for Manifold Quality}
It is difficult to find an impartial measure to use to compare manifold learning methods. If, for example, we used the cost function proposed in this paper, there would be a clear bias towards GP-MaL-MO as it has directly learned by optimising this function. Similarly, using any given measure of manifold quality is likely to be biased to some degree to some of the compared manifold learning methods. 

To avoid this issue, we instead use classification performance on the learned low-dimensional manifolds as a \textit{proxy} for measuring retained structure. Neither the proposed method nor the baseline methods use the class labels in the training/learning process (as this is unsupervised learning), and so this proxy is expected to relatively unbiased. We apply GP-MaL-MO and each of the baseline manifold learning methods to each classification dataset \textit{without} the labels present, producing a learned low-dimensional manifold. We then measure the classification accuracy attained on this transformed data by using 10-fold cross-validation with both the k-Nearest Neighbour (kNN) and Random Forest (RF) classifiers. kNN is used as an example of a very simple, distance-based classification algorithm. RF, in contrast, is much more sophisticated (using an ensemble decision-based approach) and is widely used for its high classification accuracy and applicability to a wide range of datasets \cite{zhang2017comparison}. We use the standard default implementations of this classifiers in the scikit-learn package \cite{scikit-learn}, with $k=3$ for kNN, and 100 base estimators for RF.

The \revisionOne{11} datasets used are summarised in \cref{table:datasets}. These datasets were chosen as they represent a variety of widely-used datasets, with varying numbers of features ($m$), instances ($n$), and classes ($C$). 

%\begin{itemize}
%	\item 9 different datasets, different \# features, classes, instances...
%	\item KNN and RF: explain why. One distance based/naive, one SoA common classif alg?
%	\item 10-fold cross-validation on each.
%	\item 30 GP runs on each.
%	\item GP parameters (pretty standard, note the different types of mutation).
%	\item Selection of existing MaL methods. UMAP SoA, PCA optimal for linear. (not GP-MaL --- to slow to evolve so many different \# trees!)
%\end{itemize}

\begin{table}[tb]
	%\small
	\renewcommand{\arraystretch}{1.2}
	\centering
	\caption{Classification datasets used for experiments. $m$, $n$, and $C$ represent the number of features, instances, and classes in a dataset, respectively. Most datasets are sourced from the UCI repository \cite{uci}, which contains original accreditations.}
	\label{table:datasets}
	\begin{tabularx}{0.9\textwidth}{@{}lrrrXlrrr@{}}
		\toprule
		Dataset & $m$ & $n$ & $C$ &  & Dataset & $m$ & $n$ & $C$ \\ \cmidrule{1-4} \cmidrule{6-9}
		Wine & 13 & 178 & 3 & & Madelon & 500 & 2600 & 10\\
		Vehicle & 18 & 846 & 4 &  & MFAT & 649 & 2000 & 10 \\
		Image  Segmentation & 19 & 2310 & 7 & & \revisionOne{MNIST 2-class} & 784 & 2000 & 2 \\
		Ionosphere & 34 & 351 & 2 &  & Yale & 1024 & 165 & 15 \\
		Dermatology & 34 & 358 & 6 & & COIL20 & 1024 & 1440 & 20 \\
		\revisionOne{Movement Libras} & 90 & 360 & 15 & & & & & \\
		%Iris & 3 & 150 & 3 & & Movement Libras & 90 & 360 & 15\\	
		%Breast Cancer  Wisconsin & 9 & 683 & 2 & & 
		
		%Wine & 13 & 178 & 3 & & %Isolet & 617 & 1560 & 26 \\
		 
		% MNIST 2-class & 784 & 2000 & 2\\

		\bottomrule
		
	\end{tabularx}
	%	\vspace{-1em}
\end{table}

\subsection{Experiment Settings}
GP-MaL-MO was run 30 times with different initial seeds, due to its stochastic nature. The parameter settings of GP-MaL-MO are shown in \cref{table:parameterSettings}. These parameters are reasonably standard for multi-objective GP. We used a population size of 100 as a large population is not as necessary to capture all trade-offs when one of the objectives is discrete; instead, we used a reasonably large number of generations (1000) to allow the front to be well-optimised. The ``standard'' mutation and add/remove mutation each had a 15\% probability of occurring, leaving crossover to occur the other 70\% of the time. The maximum number of trees was capped at half the number of features in the dataset. Arguably, fewer trees may be sufficient, but we wanted to test if GP-MaL-MO would be able to use a large maximum number of trees effectively, by finding a good (and smaller) number of trees to use.

	\begin{table}[tb]
	%	\captionsetup{position=top}
	%	\renewcommand{\arraystretch}{1.3}
	
	%	\footnotesize
	%\small
	\centering
	%\vspace{-.5em}
	\caption{GP-MaL-MO parameter settings.}
	\label{table:parameterSettings}
	%	\vspace{-.5em}
	\begin{tabularx}{0.9\textwidth}{ll X ll}
		
		\toprule
		Parameter& Setting && Parameter & Setting\\
		\cmidrule(r){1-2}  \cmidrule(l){4-5}
		Generations & 1000 && Population Size & 100\\
		``Standard'' Mutation & 15\%  && Add/Remove Mutation & 15\% \\
		Crossover & 70\% && Max.\ No.\ Trees & $\max(2,\lceil m \div 2 \rceil)$\\
		Min. Tree Depth & 2 && Decomposition Approach & Tchebycheff \\
		Max. Tree Depth & 14 && Pop. Initialisation & Half-and-half\\
		 
		\bottomrule
	\end{tabularx}%
	%		\captionsetup{position=bottom}
	\vspace{-1em}
\end{table}

\section{Results}
We will evaluate the performance of GP-MaL-MO in two stages: first we will discuss the approximated Pareto fronts attained by GP-MaL-MO, and how this front maps to classification accuracy; we will then compare GP-MaL-MO to the baseline classifiers.

\subsection{Multi-objective Performance of GP-MaL-MO}
\Crefrange{fig:wine-plots}{fig:COIL20-plots} show the results of GP-MaL-MO for the \revisionOne{11} datasets, ordered by increasing dimensionality. For each dataset, there are two plots. The first plot shows the mean approximated Pareto front (over the 30 runs), with the number of trees on the x-axis and the cost on the y-axis.The second plot shows the mean 10-fold cross-validation accuracy across the 30 GP-MaL-MO runs for the two classification algorithms. RF and KNN are represented by a green and blue line respectively, with dotted lines used for training performance and solid lines for test performance. The orange \revisionOne{line} represents the 10-fold test accuracy achieved by RF when using all the original features of the dataset. This represents, in some sense, the optimal performance that can be achieved only if GP-MaL-MO successfully preserves all necessary structure from the high- to low-dimensional space. The secondary (top) x-axis on the second plot shows how large the number of trees used is compared to the original feature set size. \revisionOne{The test performance of the GP-MaL results are also shown on the second plot, using an upwards/downwards facing triangle for RF and KNN, respectively.}

\paragraph{Wine:} 
\paretoFig{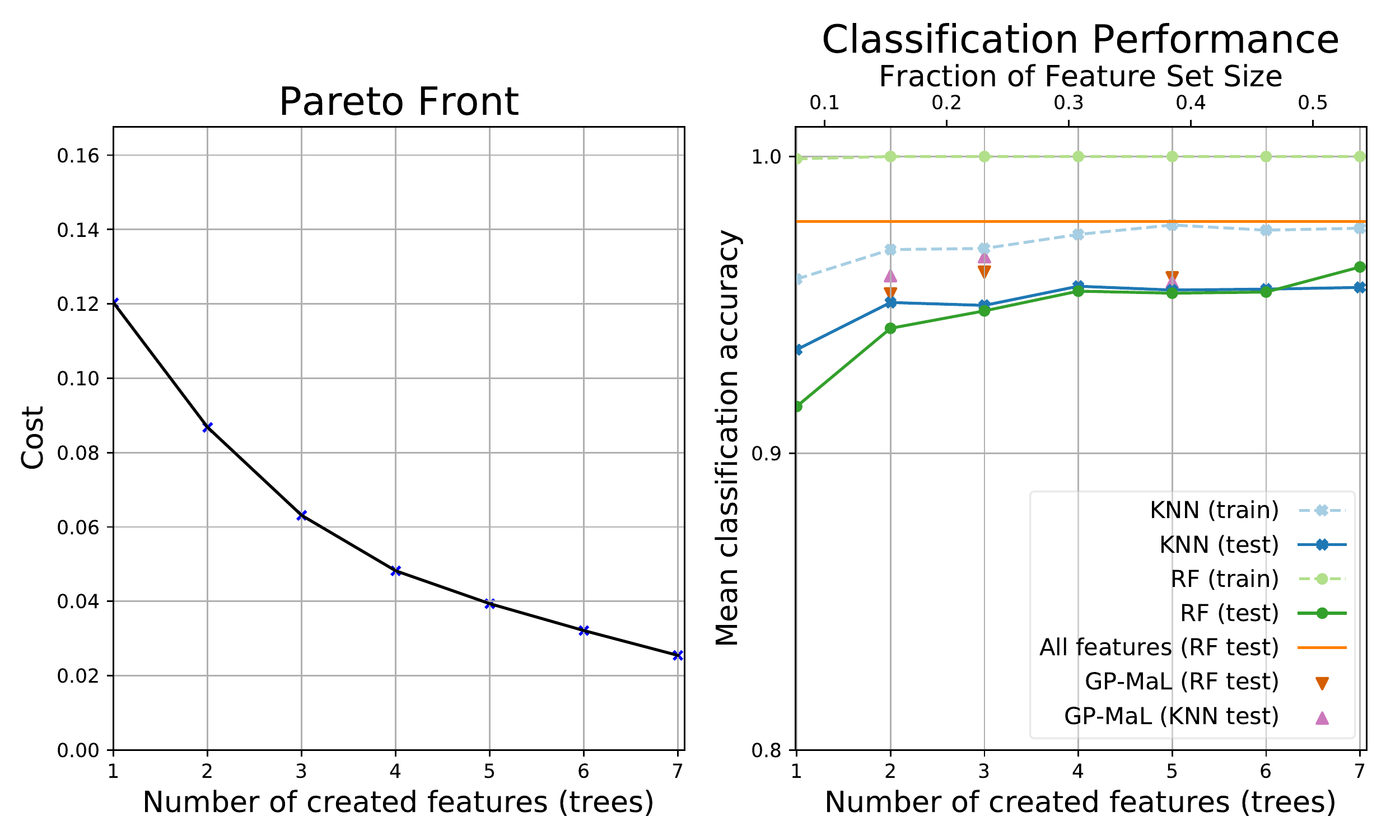}{Wine}
On the simplest dataset, GP-MaL-MO achieves over 90\% test accuracy even when only using a single tree --- the Wine dataset is quite well-separated and has a small number of features, which means one tree is enough to preserve sufficient structure. Accuracy improves slightly with additional trees, but effectively plateaus at about four features ($\approx$70\% dimensionality reduction). Compared to using all features, four created features gives only about 2\% lower RF test accuracy. The \textbf{training} accuracy of RF is nearly 100\% even with a single created feature. Given the ensemble approach of RF, it is easy for it to achieve perfect training performance --- this same pattern is seen in the remaining datasets.

\paragraph{Vehicle:}
\paretoFig{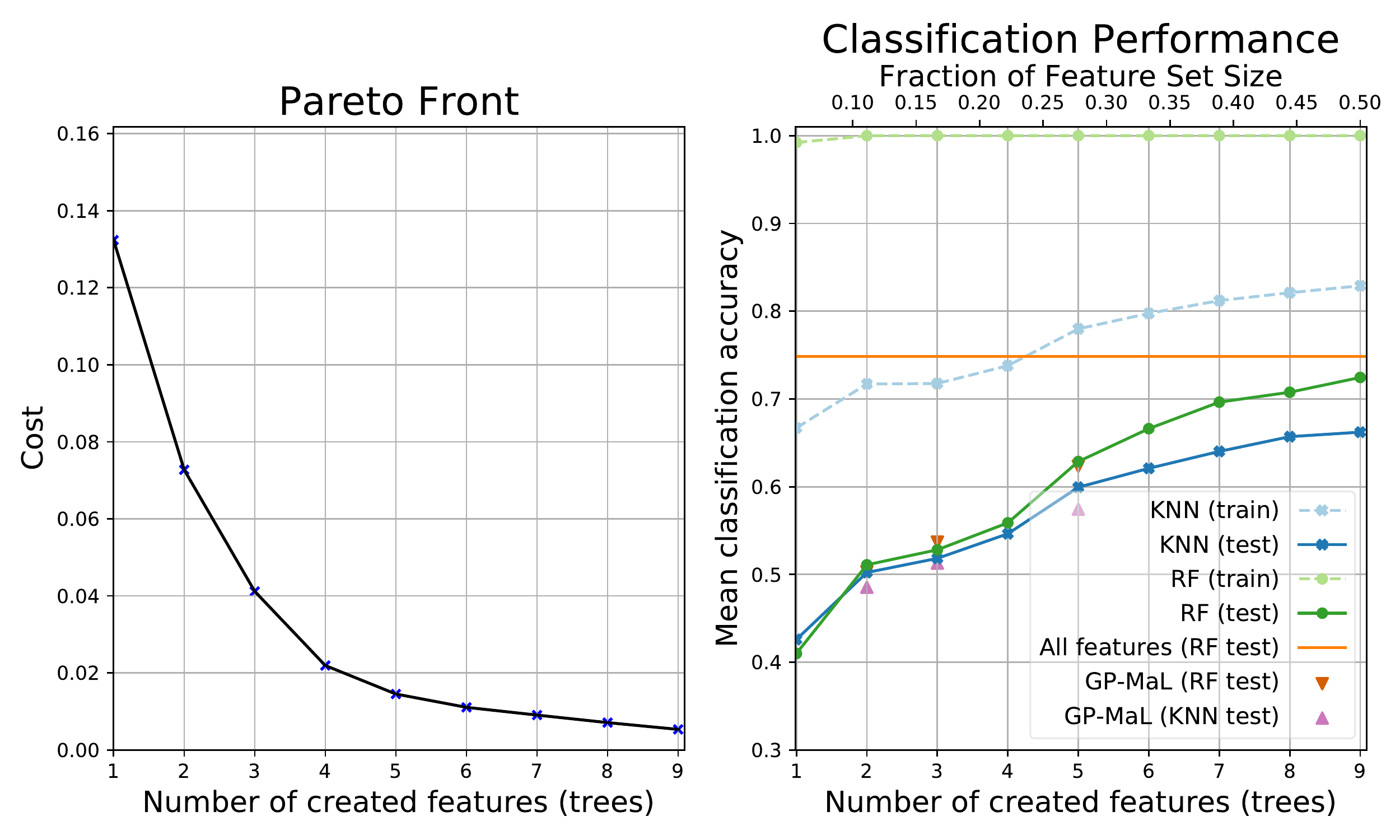}{Vehicle}
On the the Vehicle dataset, increasing the number of trees gives a much more meaningful increase in classification performance than on Wine. This is matched by large succesive drops in cost from one to four trees. Using half as many created features as original features gives a RF test accuracy within a few percentage points of using the full original feature set.

\paragraph{Image Segmentation:} 
\paretoFig{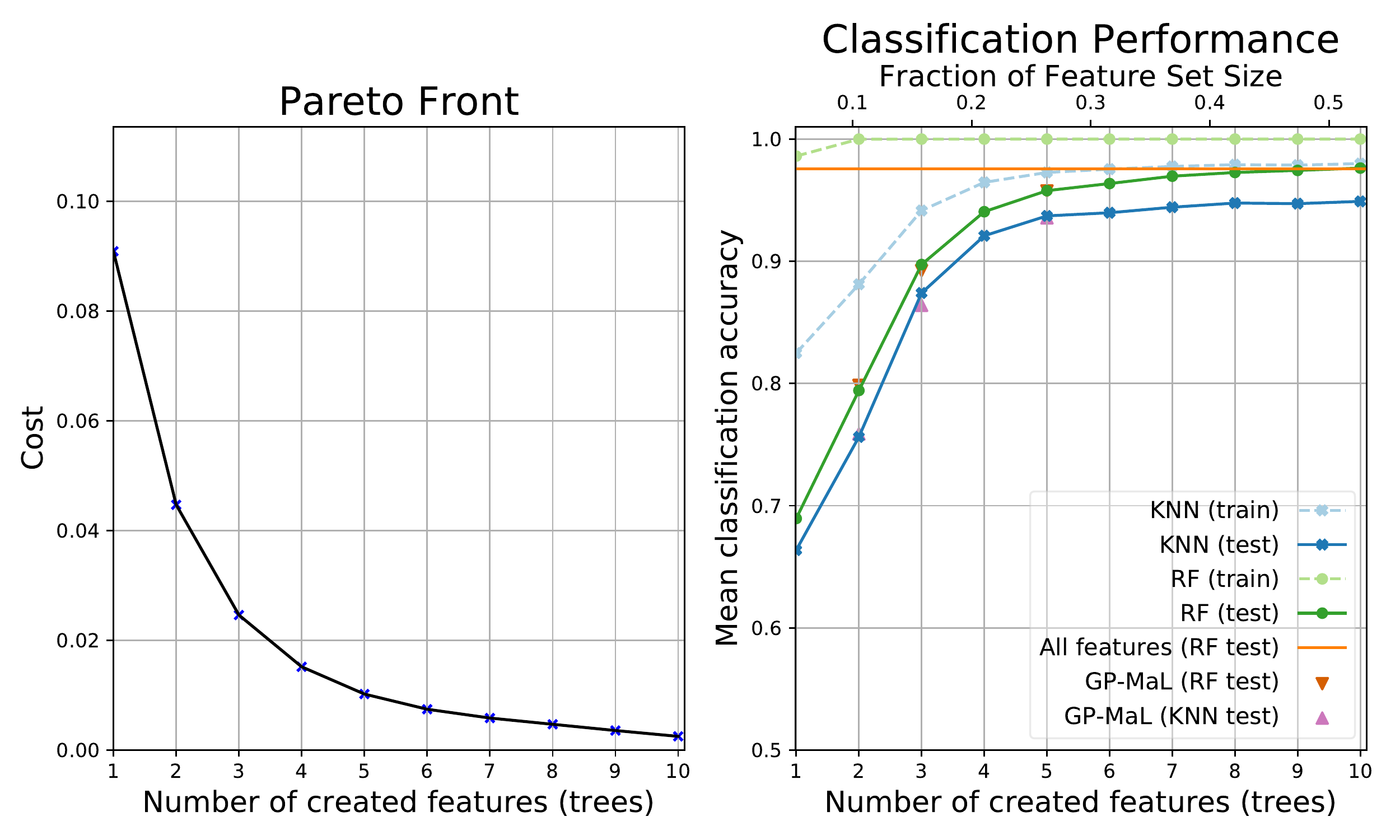}{Image Segmentation}
GP-MaL-MO produces a very similar Pareto front on this dataset as on Vehicle, but with an even stronger relationship between decreasing cost and increasing classification accuracy. As cost begins to decrease more slowly at $t=4$, test accuracy also begins to level off. By the time ten trees are used, RF test accuracy is almost identical to the accuracy achieved on the original feature set. 

\paragraph{Ionosphere:}
\paretoFig{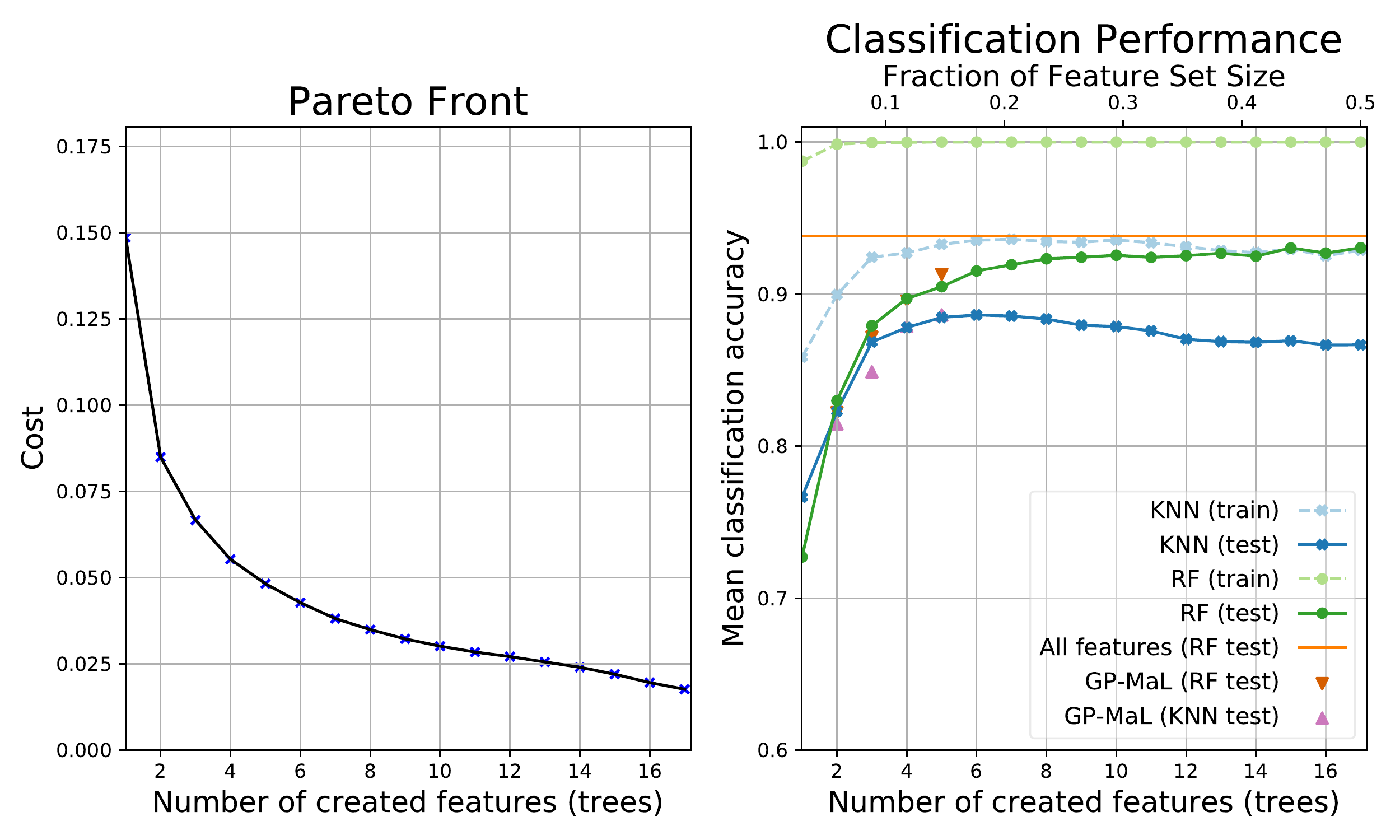}{Ionosphere}
The test accuracy for KNN on the Ionosphere dataset begins to \textbf{decrease} when more than six trees are used. RF, in contrast, continues to slowly increase in accuracy until around 13 trees. This is perhaps due to the ``naive'' approach of KNN --- by using only raw distances to perform classification, KNN may become overly sensitive at a high number of trees on this dataset. 

\paragraph{Dermatology:}
\paretoFig{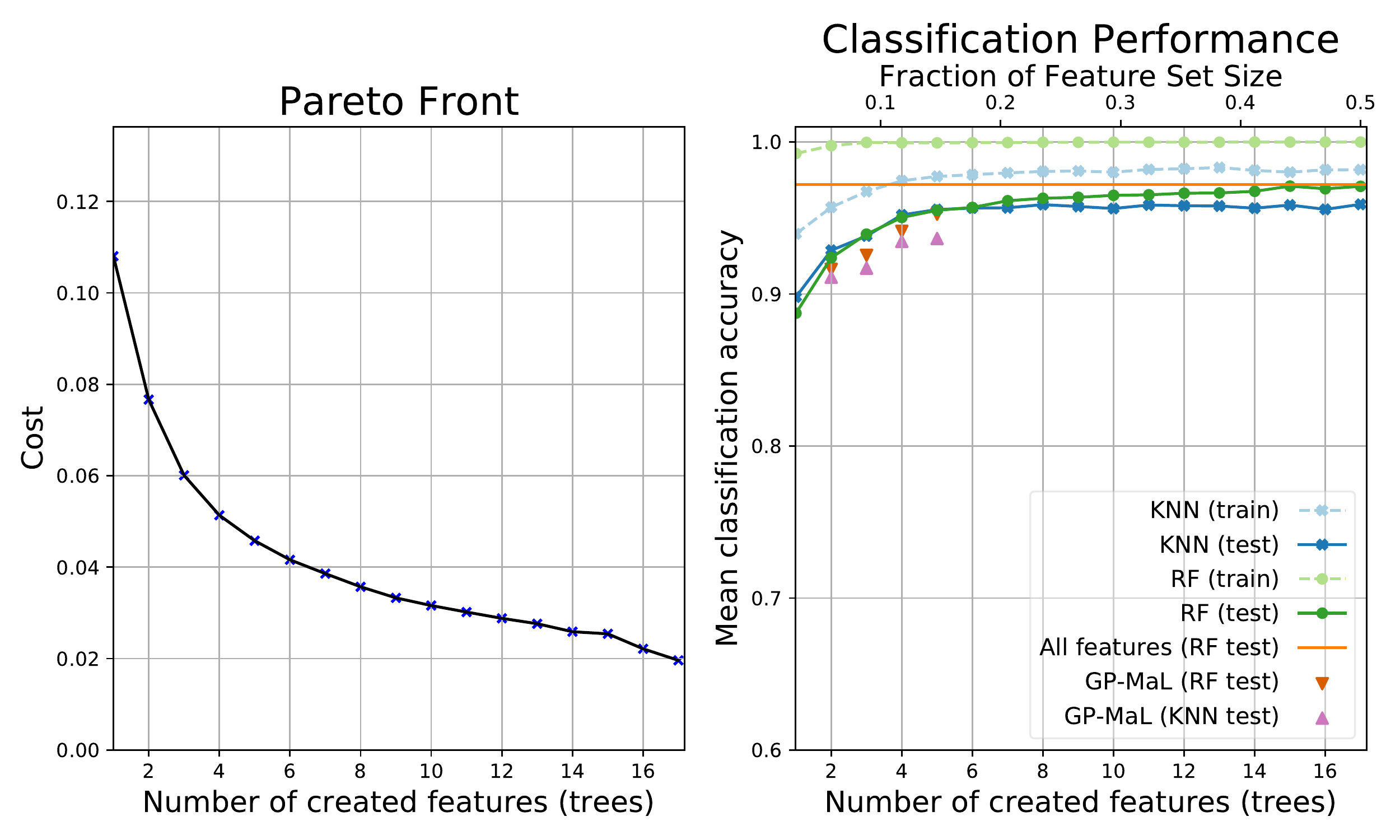}{Dermatology}
The Dermatology dataset shows a similar pattern to the Ionosphere dataset, except that both RF and KNN are able to achieve 90\% test accuracy when only a single tree is used.

\revisionOnePar{
\paragraph{Movement Libras:}
\paretoFig{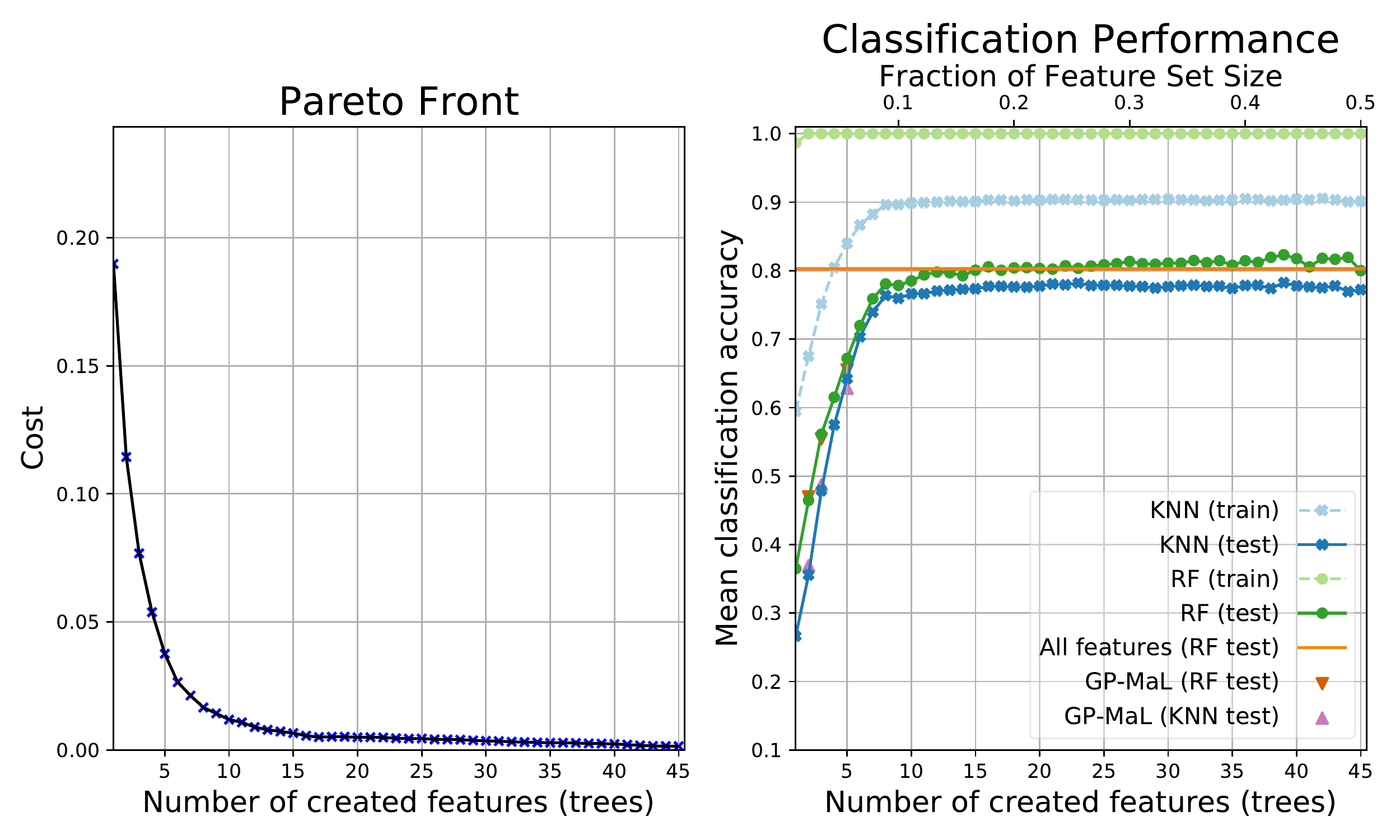}{\revisionOne{Movement Libras}}
The plots for the Movement Libras dataset show clear diminishing returns once a certain number of trees are used (t $\approx 15$). Very little further improvement is gained at higher values of $t$, which highlights one of the key benefits of GP-MaL-MO: from examining the approximate Pareto front, we can clearly see the majority of the manifold structure has been preserved by $t=15$, suggesting this is an appropriate number of created features on this dataset.
}
\paragraph{Madelon:}
\paretoFig{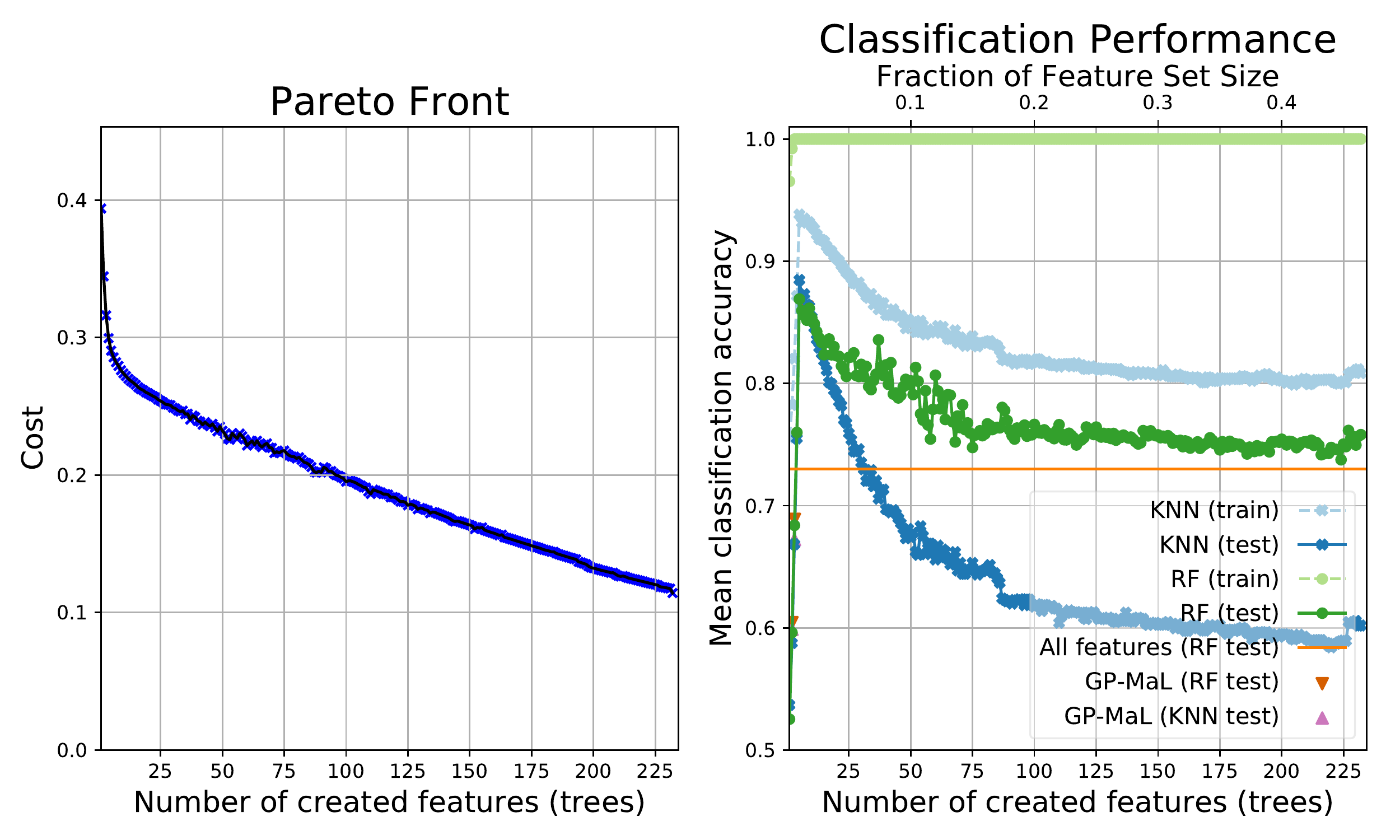}{Madelon}

The Madelon dataset produces remarkably different patterns to any of the datasets used in this paper. The approximated Pareto front does not appear as a log curve, but rather is essentially linear across most of the front. Even more notably, the classification performance on Madelon does not plateau --- rather, there is a sharp spike in test accuracy around five trees and then it quickly drops. 

Madelon is an artificial dataset designed for testing feature selection algorithms. Of its 500 features, only five are actually useful. The remaining features consist of 480 ``useless'' (noisy) features; 10 repeated features; and 5 redundant features. The spike in classification performance at five trees corresponds to GP-MaL-MO discovering the five useful features\footnote{\revisionOne{This was verified by examining the learned trees.}}, and producing a low-dimensional manifold consisting solely of them (or their redundant features). At more than five trees, GP-MaL-MO is likely attempting to preserve the ``structure'' of the noisy features, which leads to KNN and RF struggling to classify accurately. 

This pattern shows the power of GP-MaL-MO to preserve the most important structure (the useful features) of a dataset at a small number of trees --- despite there being 500 features available, GP-MaL-MO is able to reliably detect the five best ones.
\paragraph{MFAT, \revisionOne{MNIST 2-class}, Yale, and COIL20:}
{
%put them all on a page together.
\begin{figure}%[p]
	\vspace{-.5em}
	\centering
	\includegraphics[width=.865\linewidth]{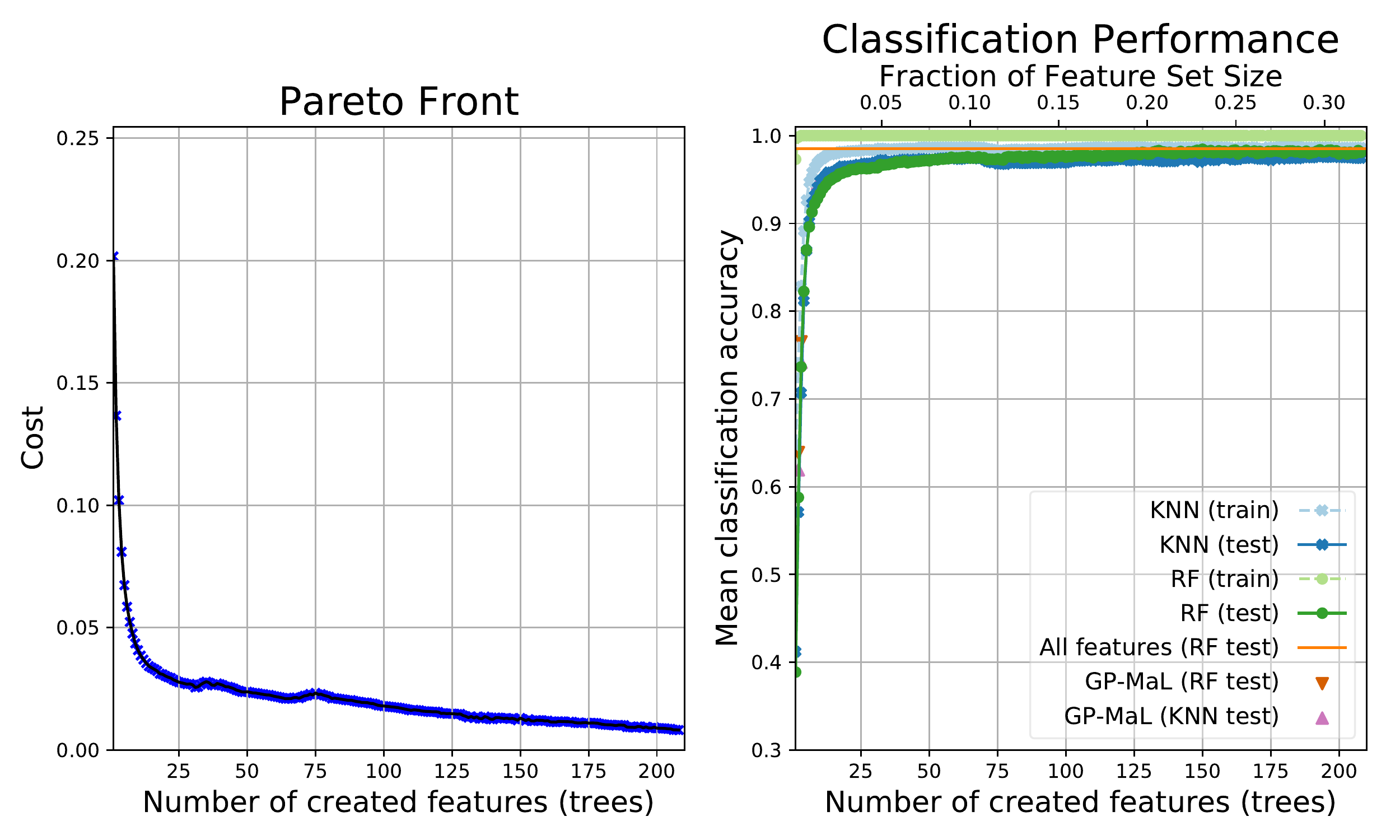}
	\vspace{-.5em}
	\caption{Approximated Pareto front on the \textit{MFAT} dataset.}
%	\vspace{-.5em}
	\label{fig:mfatCombined-plots}
%\end{figure}%
%\begin{figure}%[p]
	\vspace{-.5em}
\centering
\includegraphics[width=.865\linewidth]{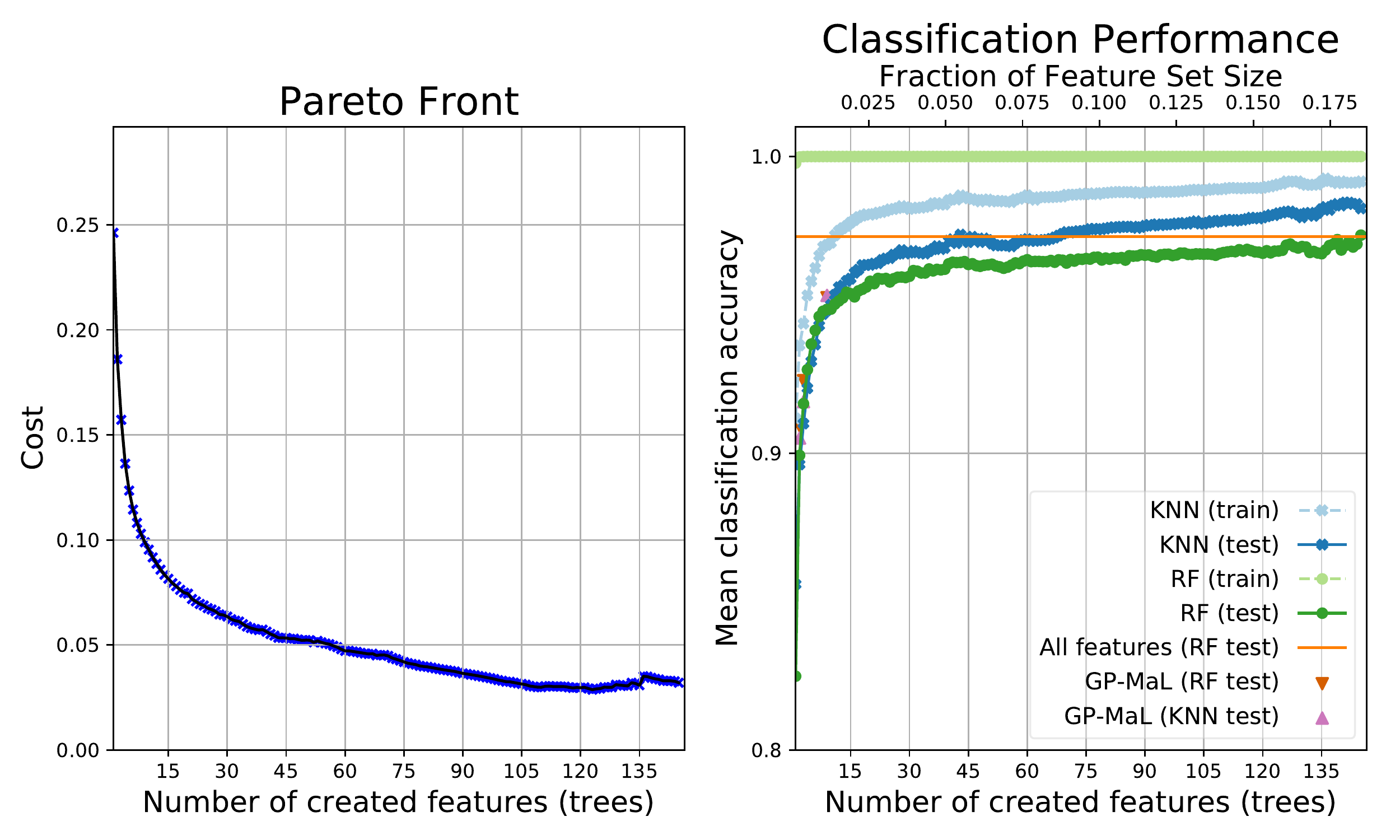}
\vspace{-.5em}
\caption{\revisionOne{Approximated Pareto front on the \textit{MNIST 2-class} dataset.}}
\vspace{-1em}
\label{fig:mnist-plots}
\end{figure}
\begin{figure}
	\vspace{-.5em}
	\centering
	\includegraphics[width=.865\linewidth]{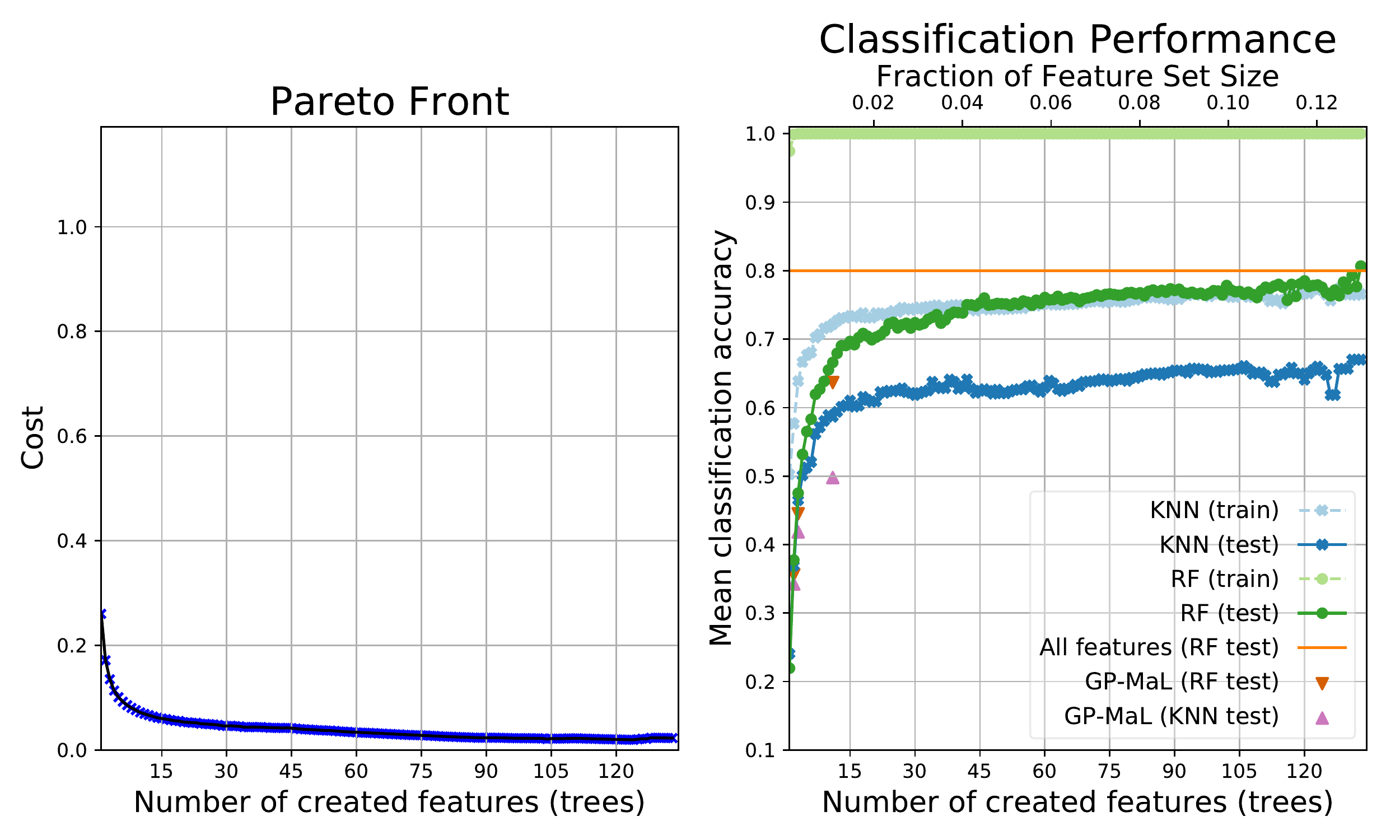}
	\vspace{-.5em}
	\caption{Approximated Pareto front on the \textit{Yale} dataset.}
%	\vspace{-1em}
	\label{fig:Yale-plots}
%\end{figure}%	
%\begin{figure}%[p]
	\vspace{-.5em}
	\centering
	\includegraphics[width=.865\linewidth]{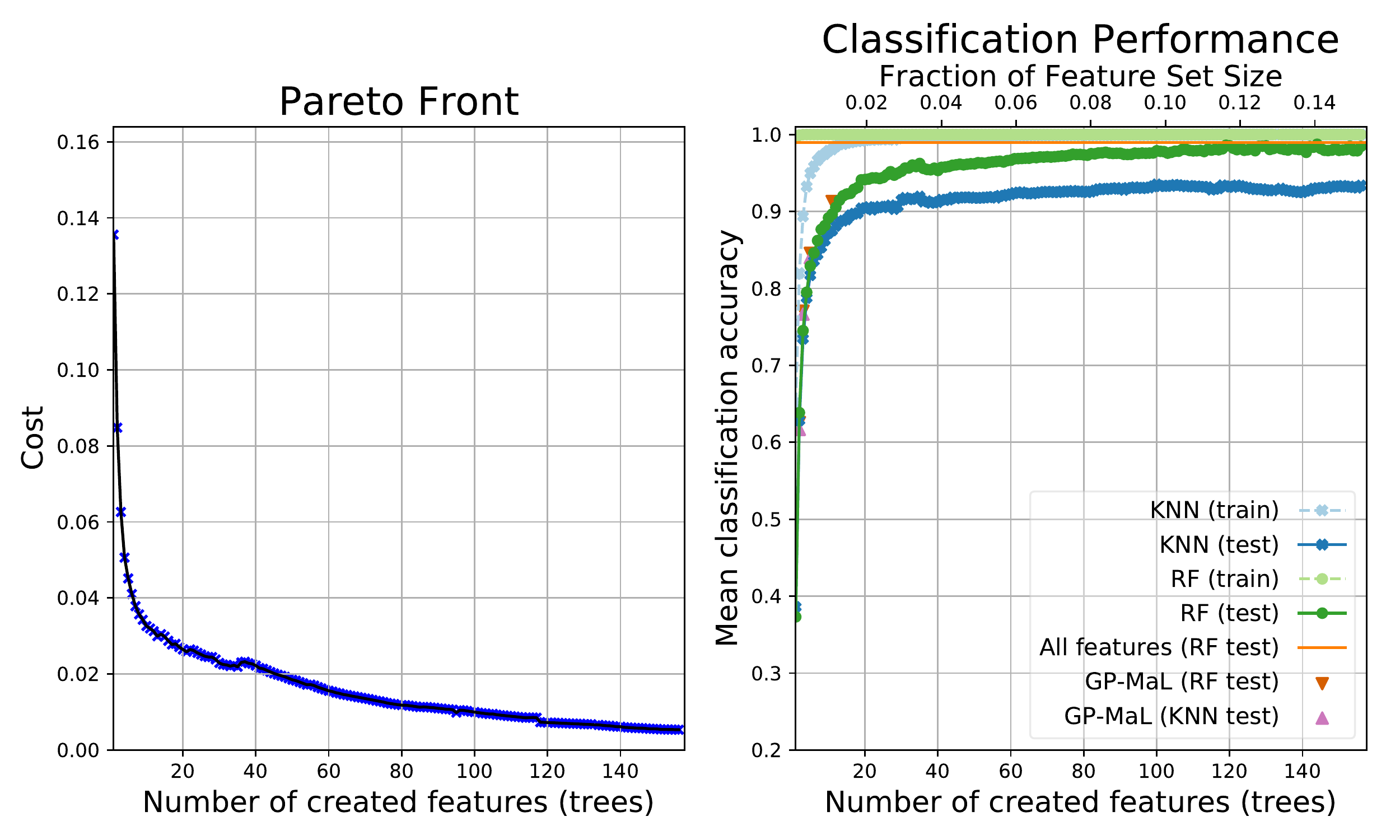}
	\vspace{-.5em}
	\caption{Approximated Pareto front on the \textit{COIL20} dataset.}
	\vspace{-1em}
	\label{fig:COIL20-plots}
\end{figure}
%	\paretoFig{mfatCombined}{MFAT}
%\paretoFig{Yale}{Yale}
%\paretoFig{COIL20}{COIL20}
}
%paragraph{Yale:}
The final \revisionOne{four} datasets, which have the highest dimensionality, all have quite similar patterns and so we will discuss them together. On all three datasets, cost decreases substantially until around 15--30 trees, at which point it decreases only very slowly as additional trees are added. On MFAT, MNIST, and COIL20, the approximated Pareto front is not entirely concave --- this is due to the front being averaged over 30 runs: not all runs produce all numbers of trees. 

On all three datasets, classification accuracy begins to plateau at around 30--45 trees. At this point, test accuracy on RF is nearly the same as when using all features in the original feature set, despite the number of trees being only about 5\% of the number of original features. On the Yale and COIL20 datasets, which have 1,024 features, GP-MaL-MO never uses more than $\approx$ 155 features, which represents a dimensionality reduction of over 80\%. This is a particular benefit of a MO approach: GP-MaL-MO eliminates individuals with higher numbers of trees as they are dominated by other individuals who have fewer trees at the same (or lower) cost.

\subsubsection{General Findings}
Across all the datasets tested (except Madelon), GP-MaL-MO was able to produce a well-formed approximate Pareto front that gives a clear trade-off between complexity (number of trees) and cost (i.e.\ approximate classification performance). There was a very strong correlation between cost and test accuracy: the test accuracy plot is essentially an inverse of the approximate Pareto front. 

Another pattern found across all the tested datasets is the diminishing returns achieved when increasing the number of trees. Above about 30--45 trees, there is very little further decrease in test accuracy as the number of trees rises. On the datasets with highest dimensionality, this means that up to a 95\% reduction in dimensionality can be achieved without much loss of manifold structure. This highlights the power of manifold learning --- by retaining maximum structure in a dataset, the data can be represented in a much smaller space without much loss in meaningful information.
%discuss classification etc.

%\paragraph{COIL20:}

%\begin{itemize}
%	\item Two plots for each dataset. Each point on a plot is the mean for all 30 of the runs which had a tree at that point. For example, if only 5 runs got 100 trees in the final Pareto front, then mean is only of 5. (This explains why averaged Pareto front is not always concave).
	%\item Left plot shows Pareto front: trade-off between cost and \# trees (lower is better). Number of trees is at most half the number of features in the dataset (but can be less if GP decides to!).
	%\item Right plot shows how classification performance increases as more and more trees are used. Reinforce that GP-MaL-MO is \textbf{unsupervised}: this classification is performed after the fact as an evaluation technique. 
	%\item Briefly discuss each in turn, point out notable patterns. E.g.\ RF almost always 100\% training; KNN train $\propto$ test. Madelon is an artificial FS dataset: explain some of patterns there.
%\end{itemize}
%\newpage

\subsection{GP-MaL-MO compared to Baselines}
The second set of results, shown in \crefrange{fig:wine-compare-plots}{fig:COIL20-compare-plots}, show how the performance of GP-MaL-MO compares to the \revisionOne{five} MaL baseline methods, using the KNN and RF classification algorithms. Each figure contains two plots (KNN and RF), which show the test accuracy achieved by each of the methods for a range of different numbers of created features. The mean classification accuracy is calculated based on 10-fold cross-validation testing. We focus only on test performance here, as it is more indicative of actual structure retained --- as seen earlier, RF can easily achieve 100\% training accuracy. We also restrict the x-axis of the plot to at most 20 created features (trees/components). This is to make the plots clearer, and more focused on the smaller x-values, where the majority of the differences between the MaL algorithms occur. At a high number of created features, the MaL methods mostly converge, as it becomes increasingly trivial to maintain all structure in the dataset. For each method in each plot, we include the maximum test accuracy in brackets in the legend (across all numbers of created features, not just up to 20) to show the maximum potential of that method. We also include an orange \revisionOne{line} in each plot that gives the test accuracy of the classifier when using all features: this represents a type of upper-bound on performance.

\paragraph{Wine:} 
%Second set of figures compares proposed method to some MaL baselines on KNN and RF. We focus only on test performance as it is more indicative of actual structure retained (as train performance can just be very fitted/overfitted...see RF earlier!). We only show the results for the first 20 created features (trees/components) as this is where the majority of the pattern is...at higher \# features, the methods tend to converge eventually anyway. Nevertheless, the brackets on the legend for each method show the maximum performance the method achieves --- i.e.\ the maximum potential. 
\compareFig{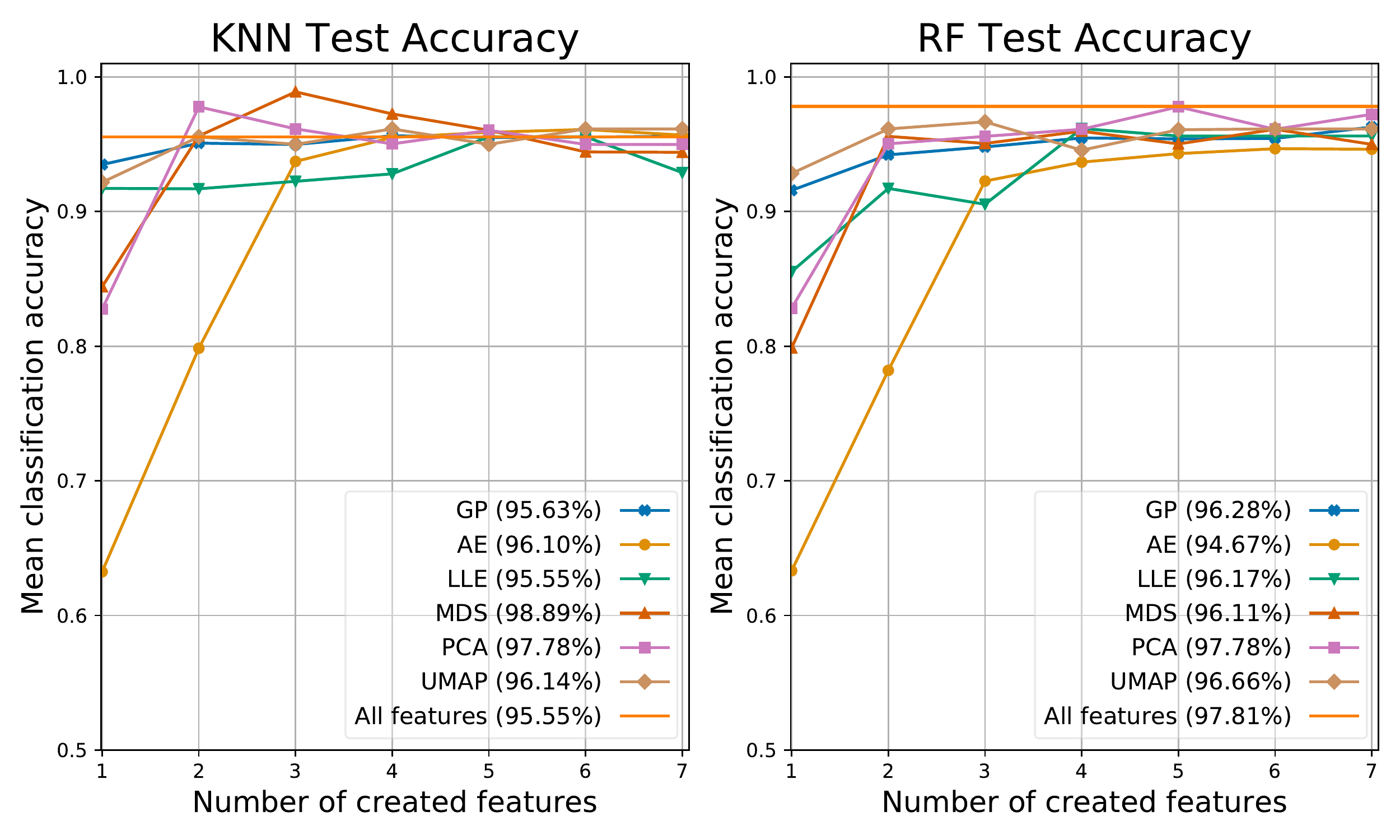}{Wine}
On the Wine dataset, the five methods give mostly comparable results, \revisionOne{except for AE, which struggles to perform competitively at a small number of created features.} On a single created feature, PCA and MDS provide about 10\% lower test accuracy using both the KNN and RF classifiers than the other MaL methods. On RF, LLE also provides clearly lower performance than UMAP and GP-MaL-MO for smaller numbers of created features.

\paragraph{Vehicle and Image Segmentation:}
\compareFig{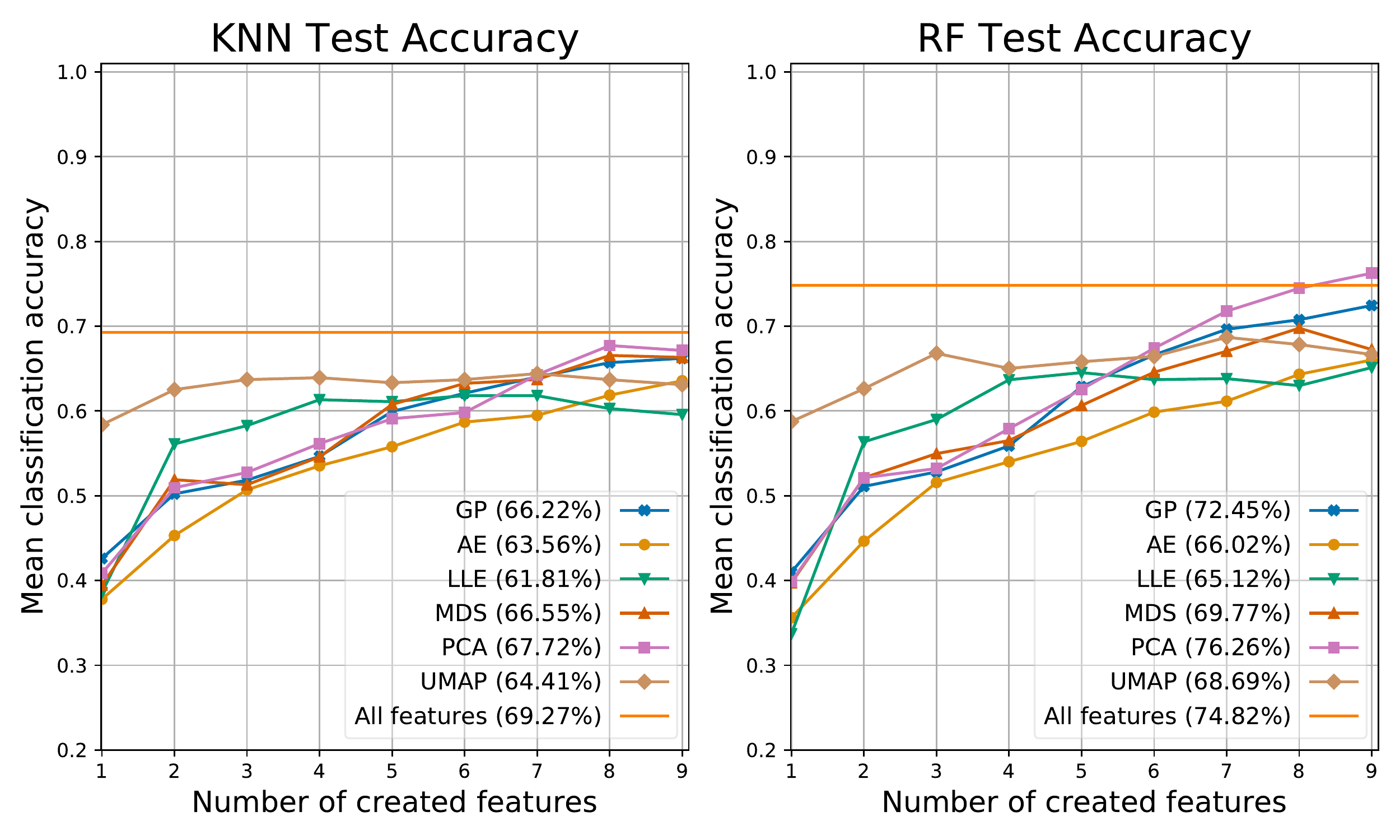}{Vehicle}
\compareFig{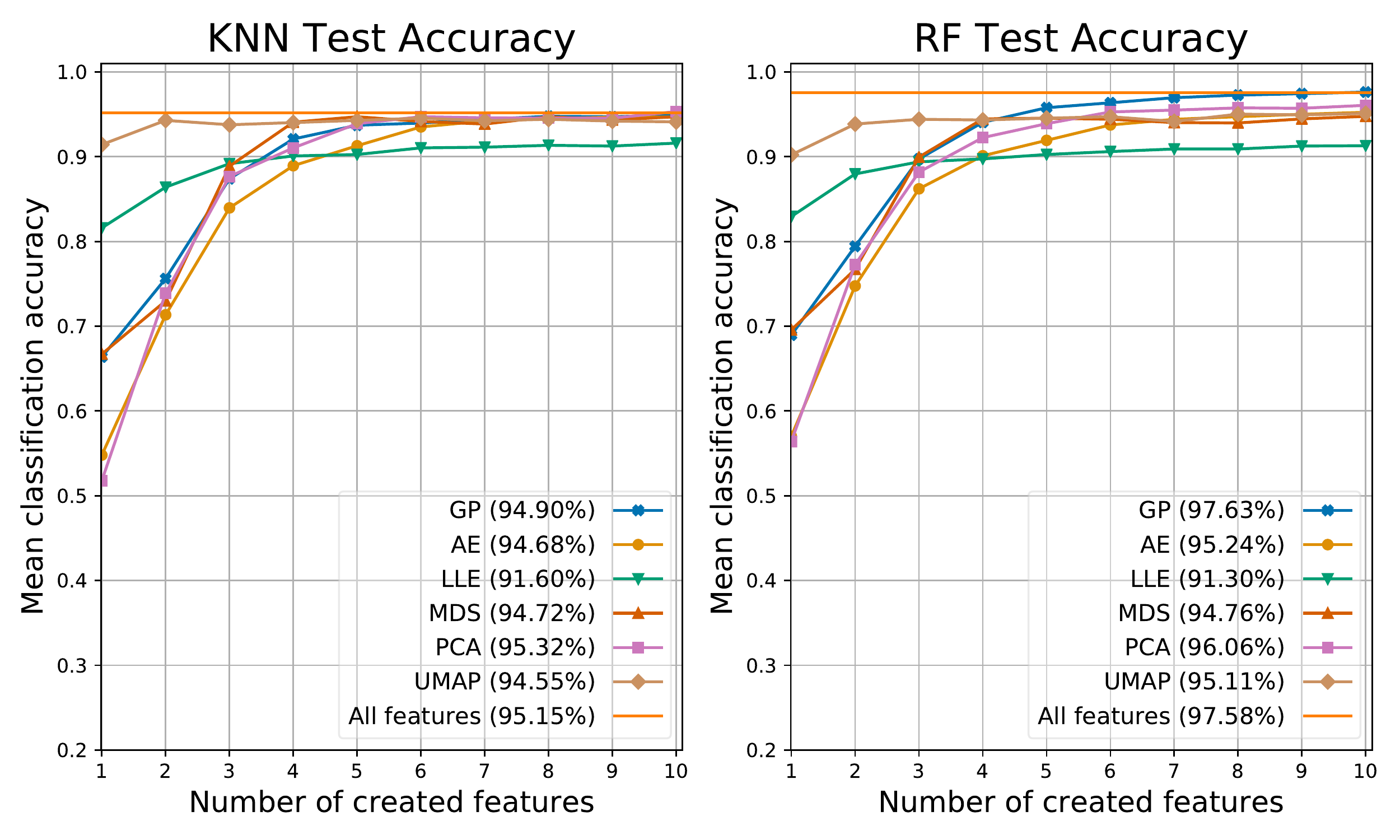}{Image Segmentation}
UMAP achieves a relatively high accuracy of $\approx$ 60\% with a single created feature on the Vehicle dataset, whereas the other methods achieve $\approx$ 40\%. By the time six created features are used, the other methods achieve similar performance to UMAP, and then GP-MaL-MO and PCA outperform UMAP at 8--9 created features. 

A similar pattern occurs on the Image Segmentation dataset, where UMAP performs very well at small numbers of created features, but is matched or slightly outperformed by other methods at higher numbers of features. GP-MaL-MO has a higher maximum accuracy than UMAP, on both these two datasets, achieving 2-3\% higher using the RF classifier. GP-MaL-MO clearly has the ability to provide high classification accuracy given enough features. The LLE method also starts out quite strongly, but has a much lower maximum accuracy than the other methods.
%\paragraph{Image Segmentation:}

\paragraph{Ionosphere:}
\compareFig{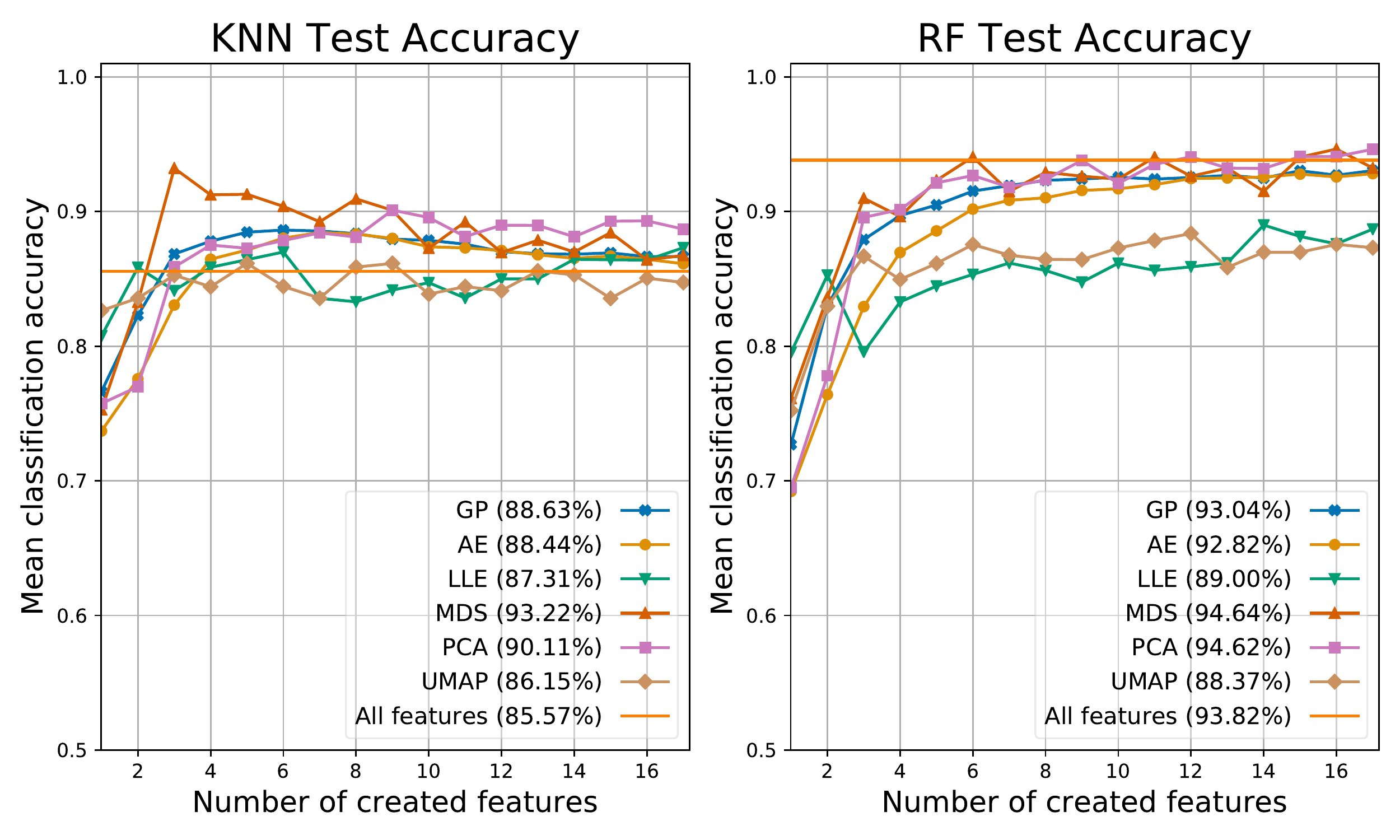}{Ionosphere}
GP-MaL-MO, PCA, \revisionOne{AE,} and MDS outperform UMAP and LLE on the Ionosphere dataset, with around 5\% higher maximum accuracy. LLE has a similar pattern to on Image Segmentation, where it has comparatively strong performance for few features, but then improves little more with additional features.

\paragraph{Dermatology:}
\compareFig{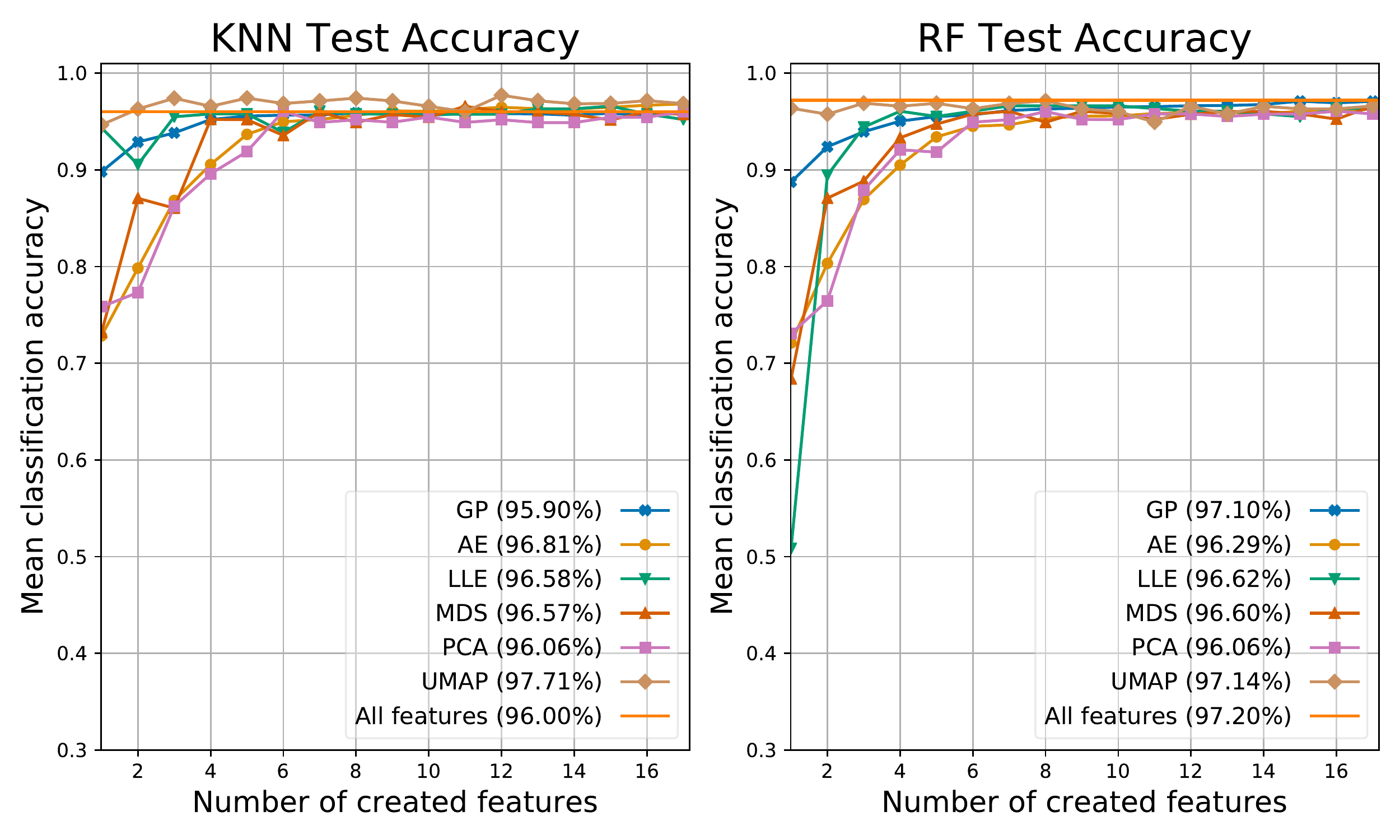}{Dermatology}
All the methods are able to achieve very good accuracy with six or more created features on the Dermatology dataset. GP-MaL-MO and UMAP both have strong performance with only a single created features, while the other methods start off worse. UMAP is clearly suited to this dataset, achieving near-maximum accuracy with a single feature.

\paragraph{Movement Libras:}
\compareFig{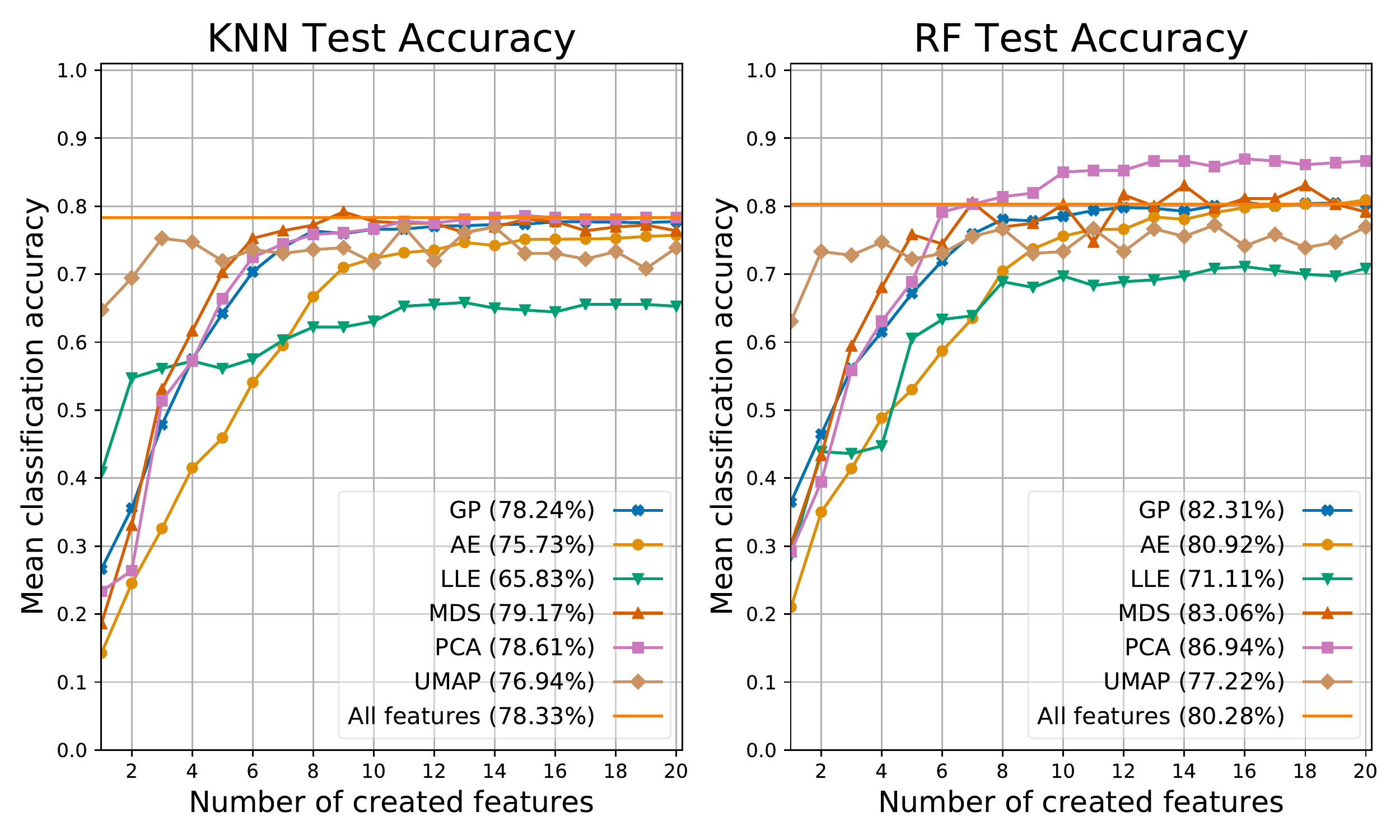}{\revisionOne{Movement Libras}}
\revisionOne{The LLE method lags behind on both classifiers on the Movement Libras dataset. UMAP again starts off with strong performance at low numbers of features, but is ultimately outperformed by GP, MDS, and/or PCA at higher dimensionality.}

\paragraph{Madelon:}
\compareFig{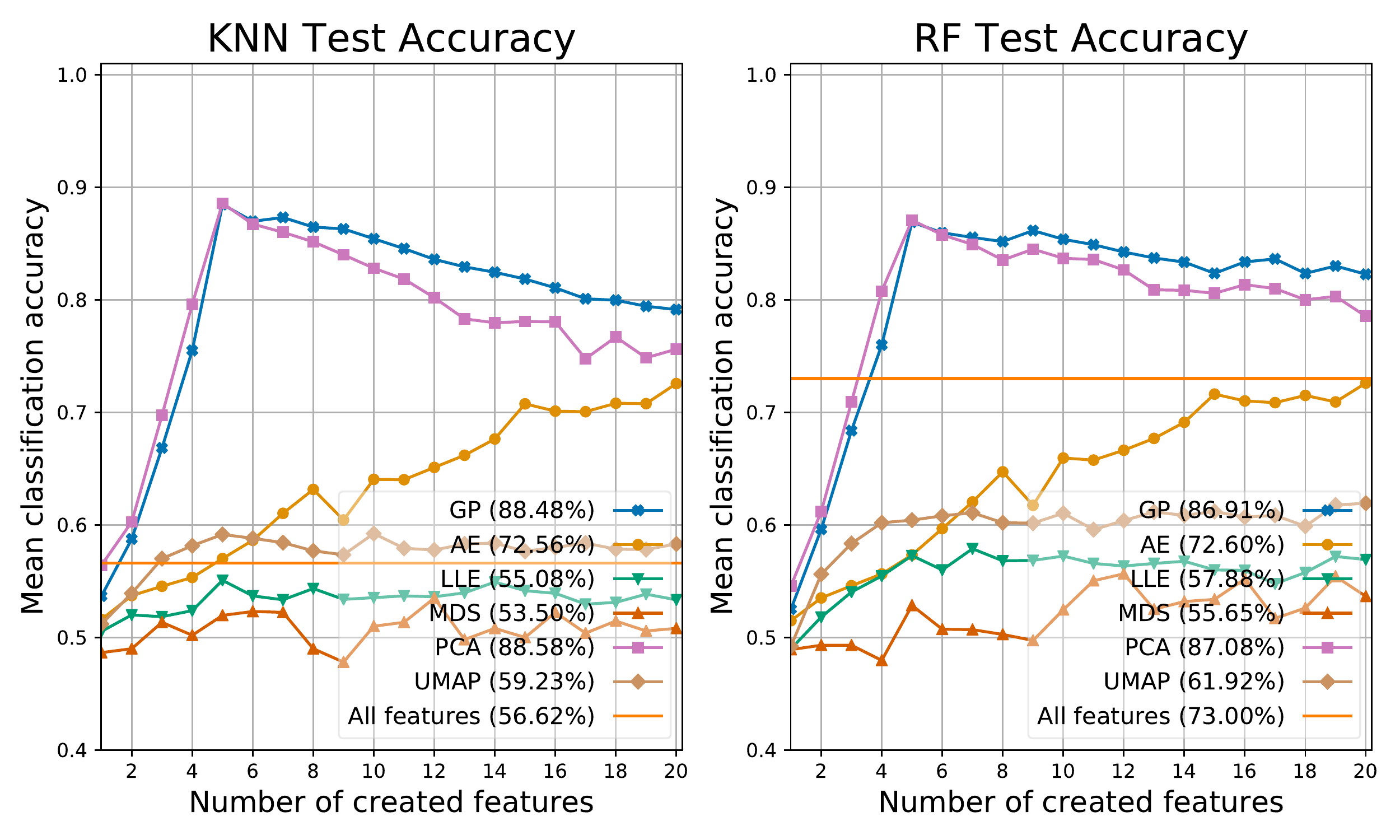}{Madelon}
The Madelon dataset again produces very distinctive patterns. Both PCA and GP-MaL-MO are able to identify the five meaningful features, whereas the other methods seem to be greatly affected by the noise in the dataset. UMAP is slightly more successful than MDS and LLE, but all three methods struggle to achieve much more than 60\% accuracy. \revisionOne{AE has poor performance when using only a few created features, but increases reasonably consistently as the number of created features rises.} It is not unexpected that PCA would perform well. PCA computes the next direction of maximum variance for each successive created feature; the directions of highest variance are very likely to be along the five meaningful features in the dataset. 

\paragraph{MFAT:}
\compareFig{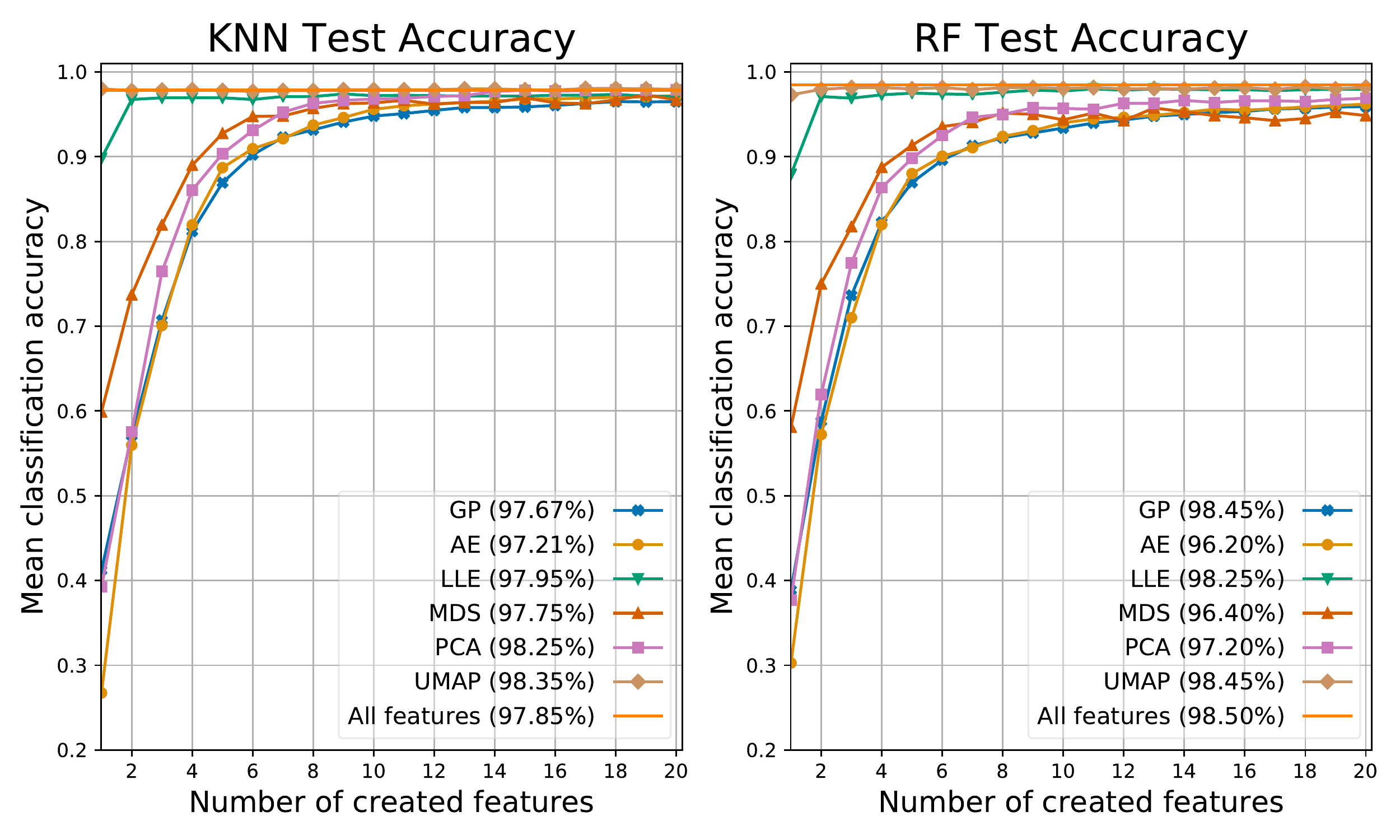}{MFAT}
The UMAP and LLE methods appear to be very well-suited to the MFAT dataset, achieving nearly 98\% accuracy with one and two created features respectively. The other \revisionOne{four} methods each follow a similar pattern, starting out with quite poor accuracy before reaching similar accuracy to UMAP and LLE at around ten created features. 

\paragraph{MNIST 2-class:}
\compareFig{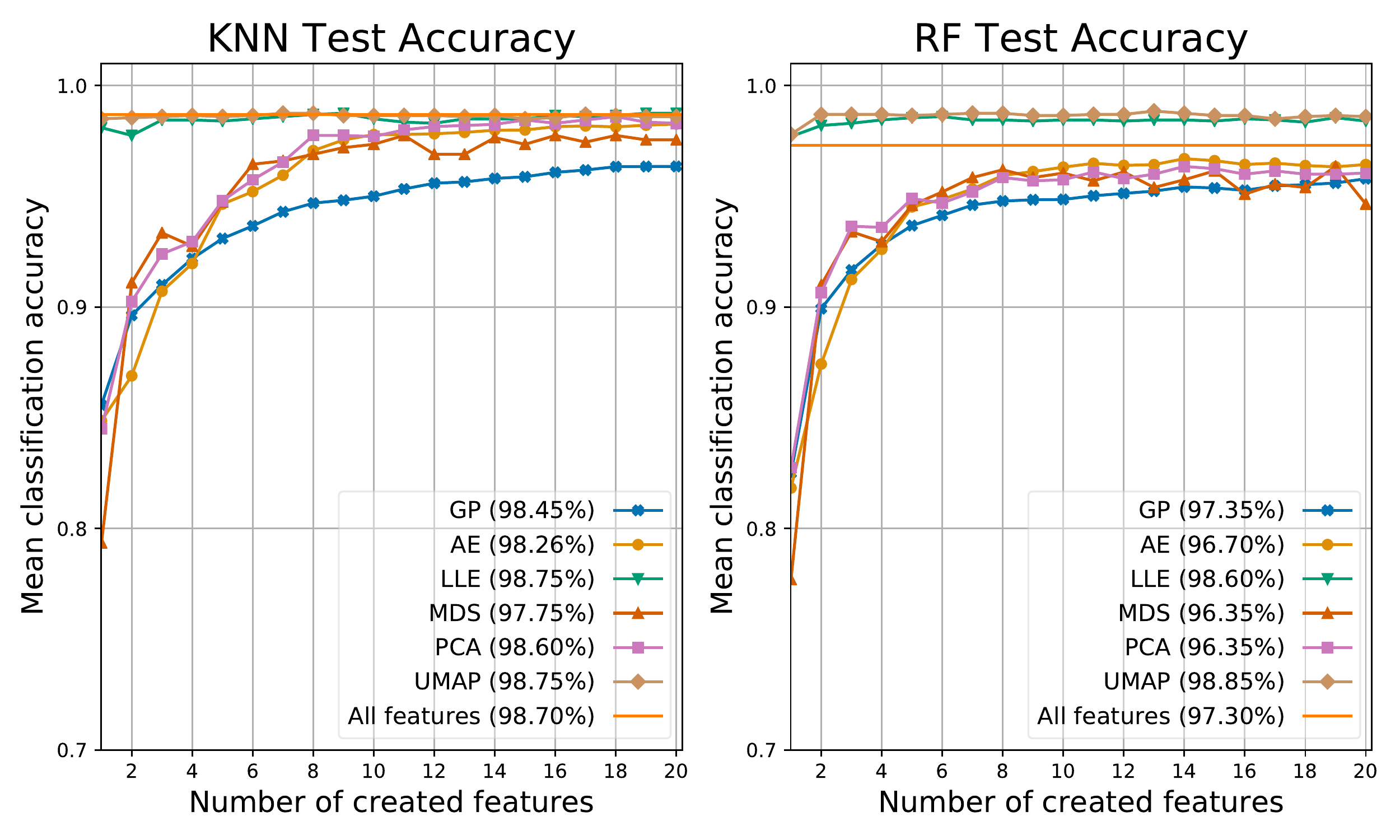}{\revisionOne{MNIST 2-class}}
\revisionOne{
As on the MFAT dataset, UMAP and LLE achieve very good results with only one or two created features. The remaining methods all achieve over 90\% classification accuracy by the time that three created features are used, and then slowly converge to around 95\% accuracy as additional constructed features are used. GP-MaL-MO appears to be slightly worse than the other methods on the KNN classifier (around 1\% less accurate), but is nearly indistinguishable when RF is used.
}
\paragraph{Yale:}
\compareFig{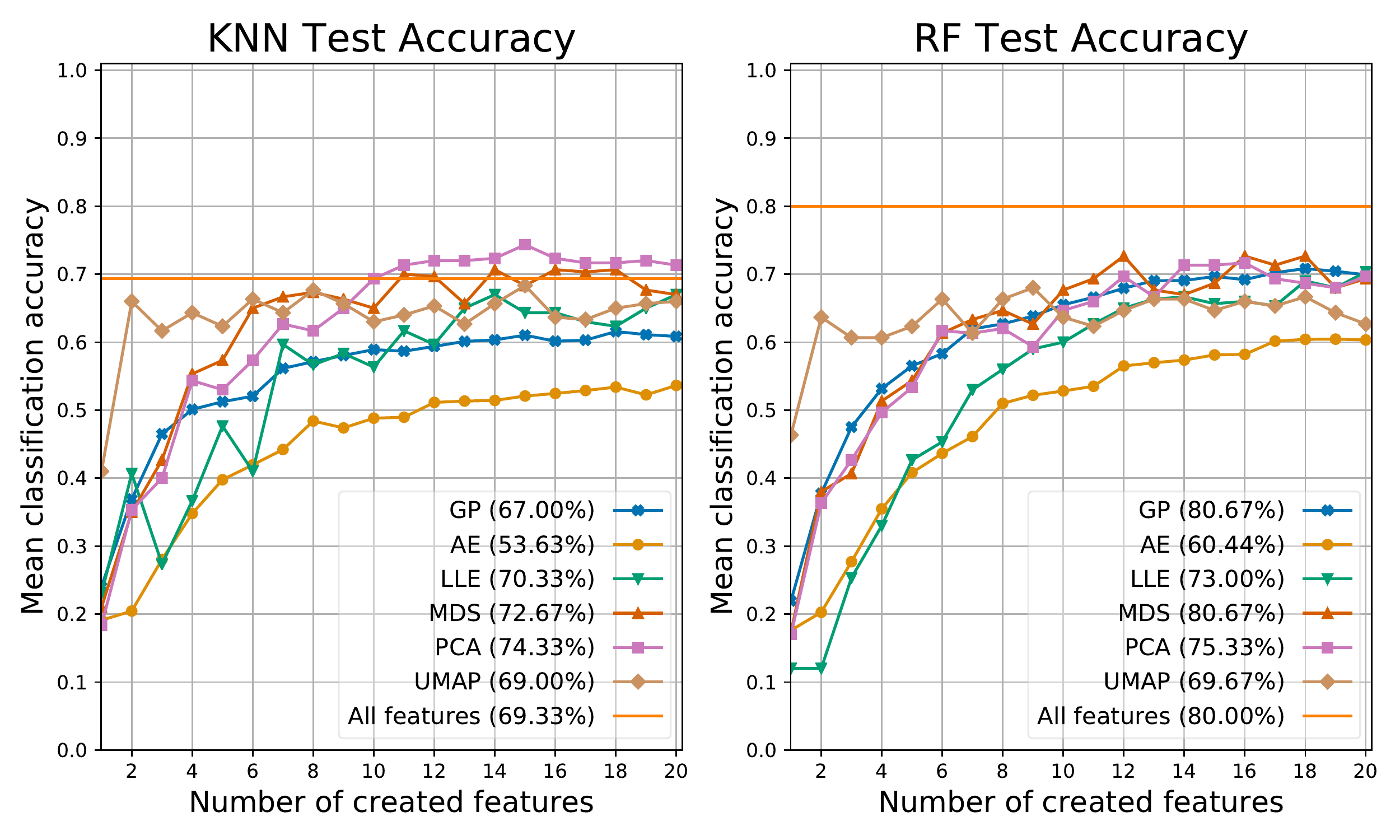}{Yale}
In contrast to MFAT and MNIST, on the Yale dataset the LLE method has arguably the worst performance at a low number of created features. UMAP again starts off quite strongly at one or two features, and then quickly levels off. With the RF classifier, the maximum performance of UMAP is 69.67\%, whereas GP-MaL-MO and MDS achieve the best accuracy of 80.67\%. GP-MaL-MO produces much better results with the RF classifier than with KNN on this dataset --- \revisionOne{this is consistent with the results achieved when using all features, which suggests that this dataset benefits from the use of a more sophisticated classification algorithm.}

%as it is only seeking to retain the ordering of neighbours (rather than the distances to them), it may be that the created features don't work as well for KNN.

\paragraph{COIL20:}
\compareFig{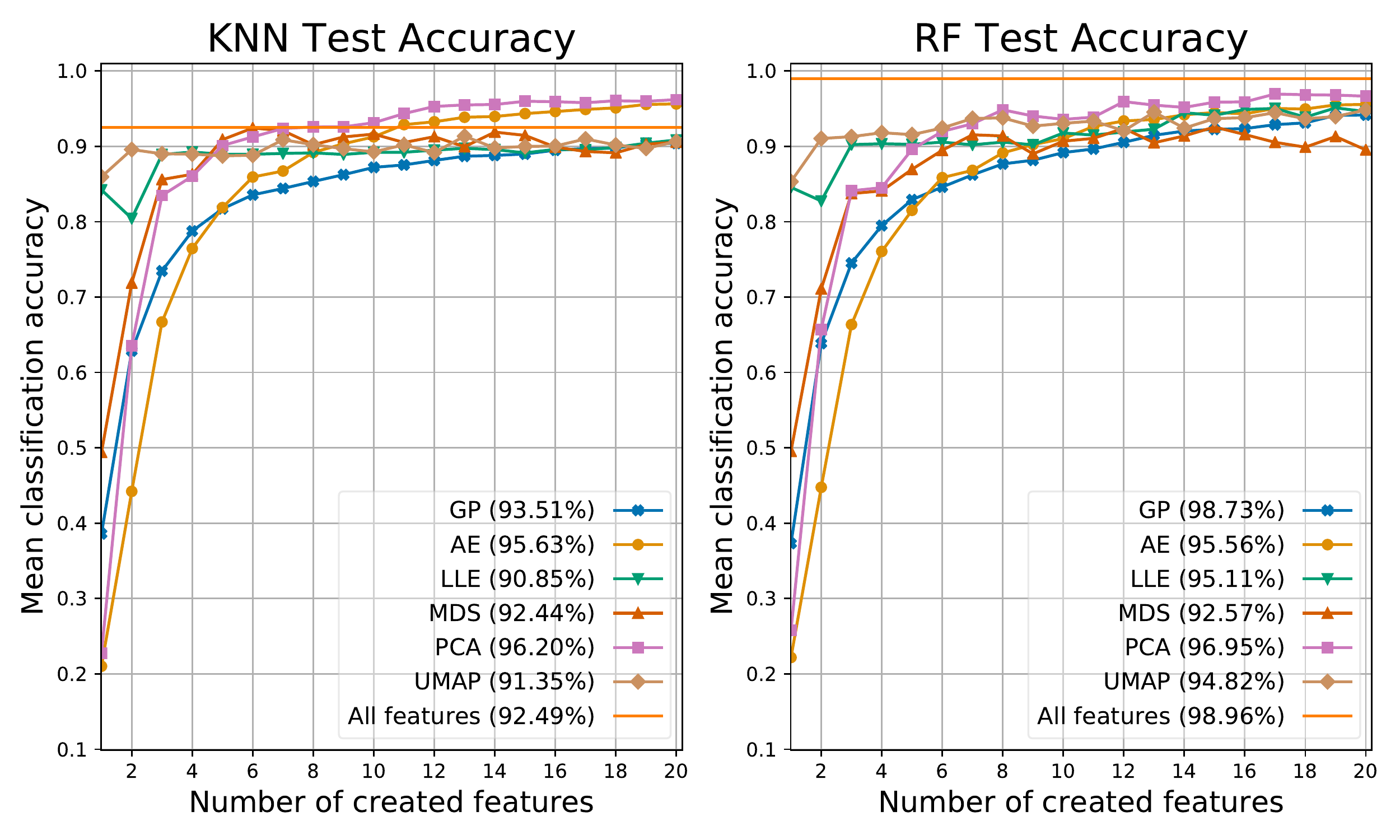}{COIL20}
On the final dataset, UMAP and LLE again both have the best performance at a small number of created features. GP-MaL-MO starts off with poor performance, but has the best overall accuracy with a sufficient number of features on RF, at 98.73\% compared to \revisionOne{95.95\% for PCA}. 

\subsection{\revisionOne{Statistical Significance Testing}}
\revisionOnePar{
To more rigorously compare GP-MaL-MO to the baseline MaL methods, we performed statistical significance testing based on the hypervolume of each of the methods on each of the two classifiers. The two objectives (\# created features and classification accuracy) were scaled to $[0,1]$ and a reference point of $[0,1]$ (representing zero classification accuracy using 20 components) was used to calculate hypervolume. The hypervolume values on the KNN and RF classifiers are shown in \cref{knnHypervolume,rfHypervolume}, respectively.

\begin{table}
\caption{Hypervolume of the MaL methods using the KNN classifier.}
\label{knnHypervolume}
\centering
\begin{tabular}{lrrrrrr}
	\toprule
	Dataset &    GP &    AE &   LLE &   MDS &   PCA &  UMAP \\
	\midrule
	wine               & 0.951 & 0.874 & 0.933 & 0.959 & 0.953 & 0.953 \\
	vehicle            & 0.564 & 0.529 & 0.576 & 0.565 & 0.564 & 0.632 \\
	image-segmentation & 0.881 & 0.852 & 0.892 & 0.885 & 0.863 & 0.941 \\
	ionosphere         & 0.873 & 0.863 & 0.864 & 0.915 & 0.874 & 0.857 \\
	dermatology        & 0.951 & 0.925 & 0.959 & 0.935 & 0.923 & 0.973 \\
	movement\_libras    & 0.685 & 0.608 & 0.612 & 0.703 & 0.687 & 0.752 \\
	madelon            & 0.833 & 0.631 & 0.544 & 0.523 & 0.839 & 0.583 \\
	mfatCombined       & 0.876 & 0.874 & 0.968 & 0.919 & 0.898 & 0.980 \\
	mnist\_train\_1k\_23  & 0.941 & 0.955 & 0.986 & 0.956 & 0.961 & 0.987 \\
	Yale               & 0.545 & 0.443 & 0.563 & 0.621 & 0.620 & 0.659 \\
	COIL20             & 0.823 & 0.833 & 0.889 & 0.883 & 0.878 & 0.904 \\
	\bottomrule
\end{tabular}
\end{table}

\begin{table}
	\caption{Hypervolume of the MaL methods using the RF classifier.}
	\label{rfHypervolume}
	\centering
\begin{tabular}{lrrrrrr}
	\toprule
	{Dataset} &    GP &    AE &   LLE &   MDS &   PCA &  UMAP \\
	\midrule
	wine               & 0.945 & 0.861 & 0.929 & 0.932 & 0.942 & 0.959 \\
	vehicle            & 0.588 & 0.534 & 0.589 & 0.582 & 0.599 & 0.657 \\
	image-segmentation & 0.907 & 0.864 & 0.893 & 0.893 & 0.878 & 0.941 \\
	ionosphere         & 0.901 & 0.884 & 0.860 & 0.918 & 0.906 & 0.866 \\
	dermatology        & 0.955 & 0.922 & 0.931 & 0.930 & 0.922 & 0.970 \\
	movement\_libras    & 0.722 & 0.658 & 0.627 & 0.746 & 0.759 & 0.754 \\
	madelon            & 0.821 & 0.641 & 0.567 & 0.534 & 0.828 & 0.600 \\
	mfatCombined       & 0.871 & 0.867 & 0.972 & 0.910 & 0.892 & 0.981 \\
	mnist\_train\_1k\_23  & 0.938 & 0.945 & 0.985 & 0.945 & 0.947 & 0.987 \\
	Yale               & 0.607 & 0.479 & 0.525 & 0.618 & 0.603 & 0.657 \\
	COIL20             & 0.843 & 0.833 & 0.915 & 0.875 & 0.887 & 0.931 \\
	\bottomrule
\end{tabular}
\end{table}

To analyse these results, we followed the methods suggested by Demsar \cite{demsar2006statistical}: we first performed a (non-parametric) Friedman's test, which found there was a difference in the mean hypervolume between the different MaL methods at a 95\% confidence level for both classifiers. The rankings produced by this Friedman's test are shown in \cref{rankingKNN,rankingRF}. Based on this, we performed a post-hoc analysis to compare the proposed GP-MaL-MO approach to each of the five baselines using a Holm test, the results of which are shown in \cref{posthoc}. The only statistically significant difference found (at a 95\% confidence level) was that UMAP out-performed GP-MaL-MO when using the KNN classifier. Given that UMAP is the state-of-the-art in MaL, and performs embedding rather than mapping, this can be seen as a positive result. No significant difference was found between UMAP and GP-MaL-MO on RF, nor between GP-MaL-MO and any other MaL method using either classifier. The fact that GP-MaL-MO is competitive with all existing MaL methods is a very promising sign for future work, especially since it produces a mapping and also provides a range of solutions representing different levels of dimensionality reduction.

\begin{table}
	\centering
	\caption{Method rankings based on the Friedman's test using the KNN classifier.}
	\label{rankingKNN}
	\begin{tabular}{lllllll}
		\toprule
		Method &  UMAP &   MDS &   LLE &   PCA &    GP &    AE \\
		Ranking & 2.279 & 3.647 & 3.989 & 4.331 & 5.356 & 6.724 \\
		\bottomrule
	\end{tabular}

	\caption{Method rankings based on the Friedman's test using the RF classifier.}
	\label{rankingRF}
	\begin{tabular}{lllllll}
		\toprule
		Method &  UMAP &   PCA &    GP &   MDS &   LLE &    AE \\
		Ranking & 2.165 & 3.761 & 4.331 & 4.672 & 4.786 & 6.610 \\
		\bottomrule
	\end{tabular}

\end{table}
\begin{table}
	\caption{Post-hoc testing of GP-MaL-MO vs the baseline methods.}
	\label{posthoc}
		\hfil
	\begin{tabular}{lrl}
		\toprule
		{KNN Results} &     p &    sig \\
		\midrule
		\textbf{GP vs UMAP} & 0.010 &   True \\
		\textbf{GP vs MDS } & 0.350 &  False \\
		\textbf{GP vs LLE } & 0.514 &  False \\
		\textbf{GP vs AE  } & 0.514 &  False \\
		\textbf{GP vs PCA } & 0.514 &  False \\
		\bottomrule
	\end{tabular}
\hfil
\begin{tabular}{lrl}
	\toprule
	{RF results} &     p &    sig \\
	\midrule
	\textbf{GP vs AE  } & 0.113 &  False \\
	\textbf{GP vs UMAP} & 0.121 &  False \\
	\textbf{GP vs PCA } & 1.000 &  False \\
	\textbf{GP vs LLE } & 1.000 &  False \\
	\textbf{GP vs MDS } & 1.000 &  False \\
	\bottomrule
\end{tabular}
	\hfil
\end{table}
}
\subsection{Discussion}
There are a number of common patterns across the \revisionOne{11} datasets, which provide a number of interesting findings and suggestions for future work.

The proposed method, GP-MaL-MO, is always able to achieve near-maximal test accuracy given a sufficent number of features, using the RF classifier. Other methods such as UMAP and LLE often have high relative performance for a small number of features, but then plateau as the number of created features increases. On most datasets, PCA and MDS provide quite similar results to GP-MaL-MO. However, PCA necessarily uses all the original features in each constructed feature; MDS provides only a low-dimensional embedding and not a way of mapping this to the original feature space. \revisionOne{While AE does produce a mapping, it has the worst performance of the MaL methods across nearly all of the datasets.}
%In the case of PCA, this makes the constructed features complex, whereas for MDS there is no way of re-using the embedding on future examples or understanding the created features in terms of the original features.

The ability of GP-MaL-MO to produce a range of models with different numbers of dimensions in a single learning process sets it apart from the other manifold learning methods. While PCA also achieves this, it does so by weighting all features for every created feature --- by using only some features for each created feature, GP-MaL-MO has the potential to create simpler and more efficient models. For UMAP, LLE, and MDS, one must re-run the manifold learning algorithm for every number of created features separately. In addition, all of these methods produce only a low-dimensional embedding, rather than a mapping between the high- and low-dimensional spaces. As such, these methods cannot be used on new instances without being re-run, and cannot be easily analysed to understand the meaning of the low-dimensional structure in terms of the original features.

The cost function used in this work (\cref{eqn:cost}) appears to be very appropriate, given that as it is minimised, classification test accuracy is maximised proportionally. Indeed, when GP-MaL-MO achieves a cost value below about 0.01, the test accuracy is very similar to that of using all features. The major limitation of GP-MaL-MO is its relatively high cost/lower accuracy compared to methods such as UMAP and LLE at a low number of features. Additional work is needed to ``push'' the approximated Pareto front further towards the origin for small numbers of features, as this will ultimately allow GP-MaL-MO to be very competitive with existing manifold learning methods across all levels of dimensionality reduction. Revising the EMO approach used to focus more on this part of the Pareto front would likely be a promising future area of research. For example, a variation of MOEA/D that decomposed the Pareto front into a larger number of vectors at smaller number of trees, or other types of evolutionary pressure would be interesting future directions to investigate.

%UMAP embedding vs mapping.

%Discuss results in turn. GP-MaL-MO struggles somewhat more on higher-dimensional datasets, but still able to match performance of best methods with enough trees. Interesting that UMAP fails on some datasets quite spectacularly. PCA annoyingly good at sufficient dimensionality --- though probably makes sense, given sufficient linear combinations of features will cover majority of variance/structure in data.
%\newpage

%\paretoFig{iris}
%\paretoFig{isolet}

%most methods can't produce multiple \#s of components at once except PCA right? Maybe fewer unique features than PCA (or has potential to). --> lead into further analysis

%fitness function clearly suitable, given classif acc increases almost monotonically, and reaches potential of all features.

%stability 

%more focus needed to ``lift up'' the very small \# of trees?

%Can we combine many trees into a single one as a post-processing style of GP? Or even just basic arithmetic.

%\paretoFig{\detokenize{mnist_train_1k_23}}

\section{Further Analysis}
%What else would be good/achievable in time-frame. Or is this unnecessary? Trying to ``understand'' trees is probably going to be hard unless there's enough of them...
To better understand the potential of GP-MaL-MO for performing manifold learning, we will analyse some sample learned individuals in this section. 

\begin{figure}[p]
\renewcommand\thesubfigure{\arabic{subfigure}}

	\begin{subfigure}[b]{.64\linewidth}
		\hspace{-.5cm}
		\includegraphics[width=\linewidth]{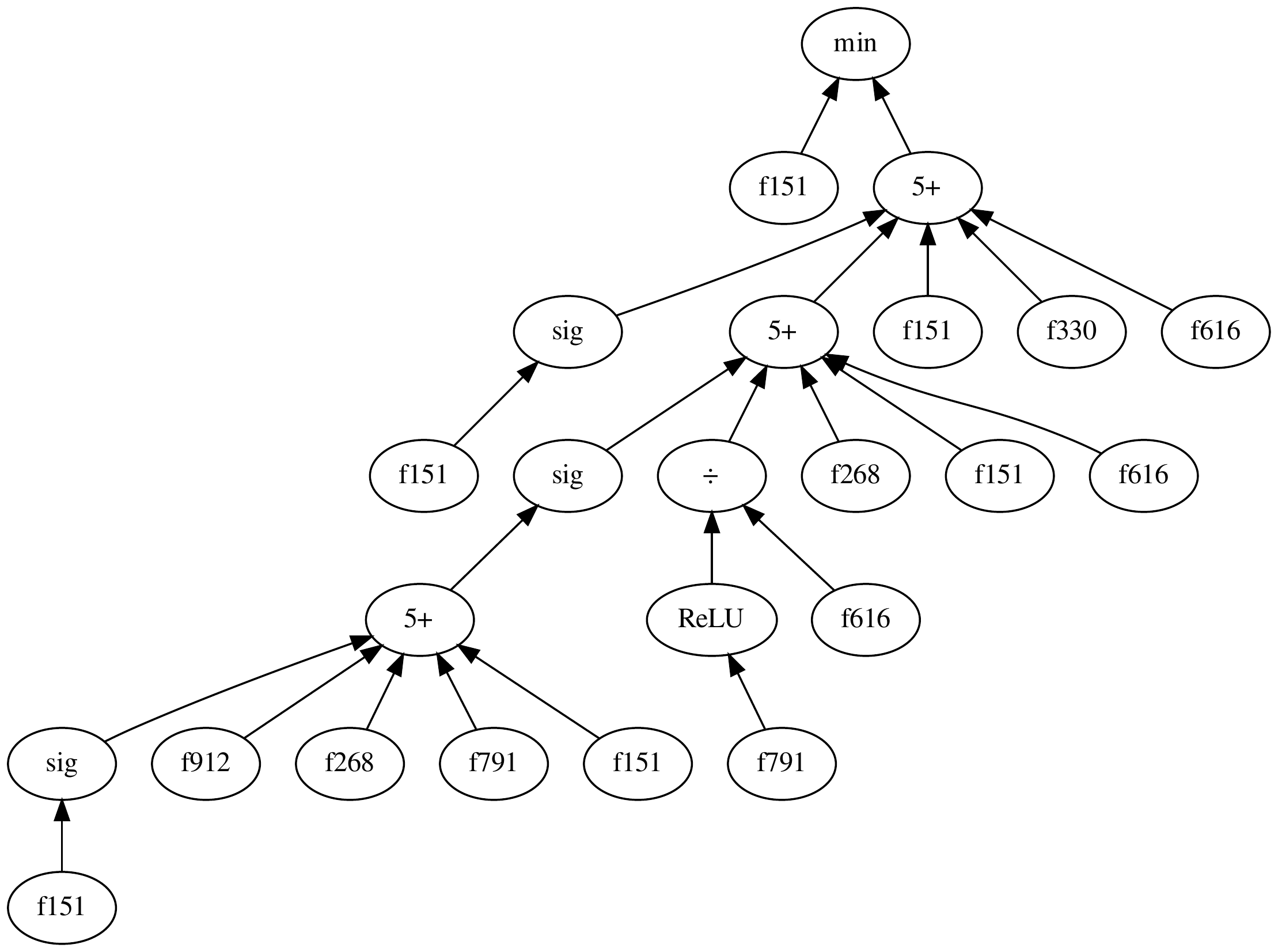}
		\subcaption{}
	\end{subfigure}%
	\begin{subfigure}[b]{.44\linewidth}
				\hspace{-.5cm}
	\includegraphics[width=\linewidth]{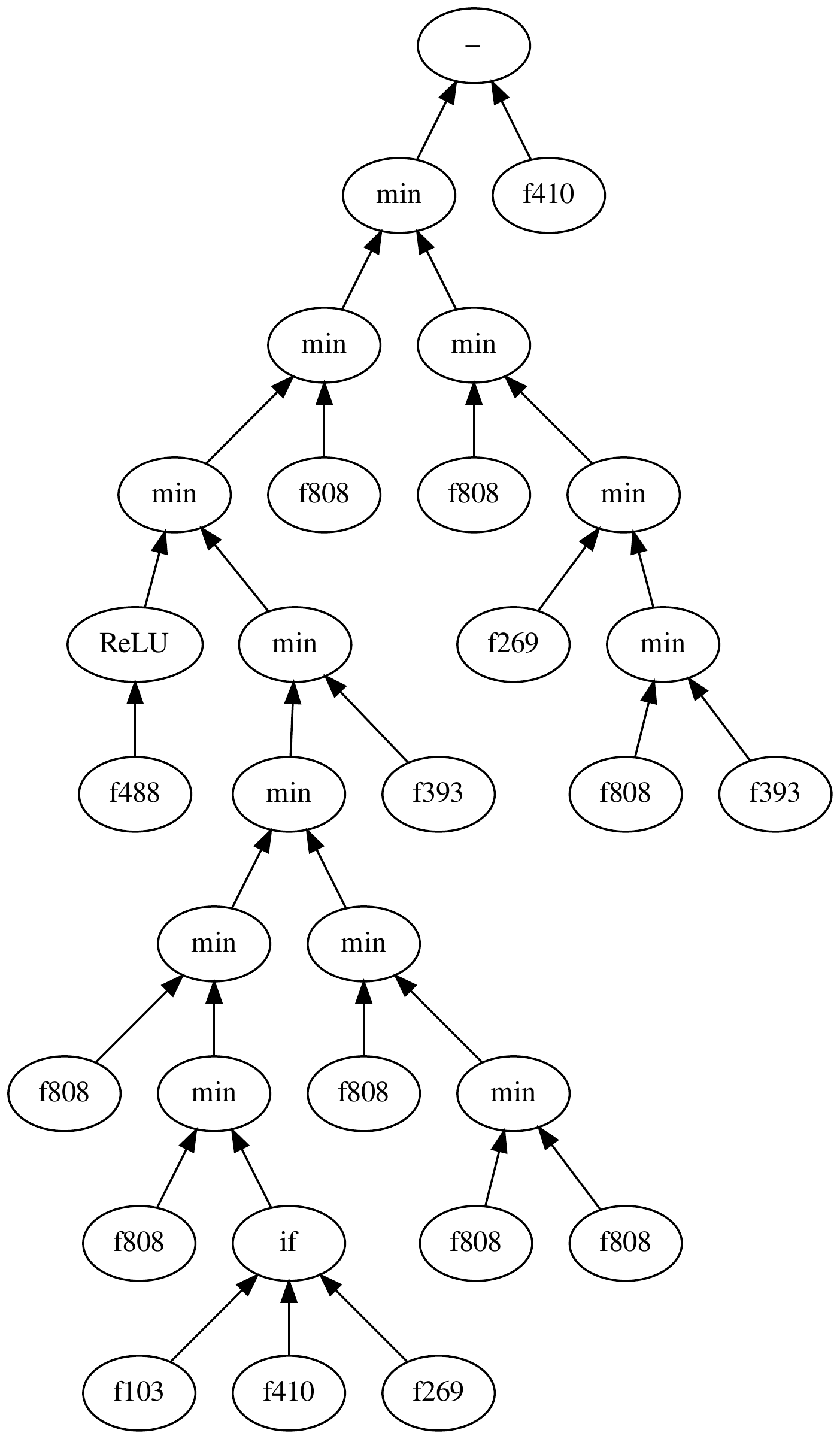}
	\subcaption{}
\end{subfigure}
	\begin{subfigure}[b]{.15\linewidth}
	\includegraphics[width=\linewidth]{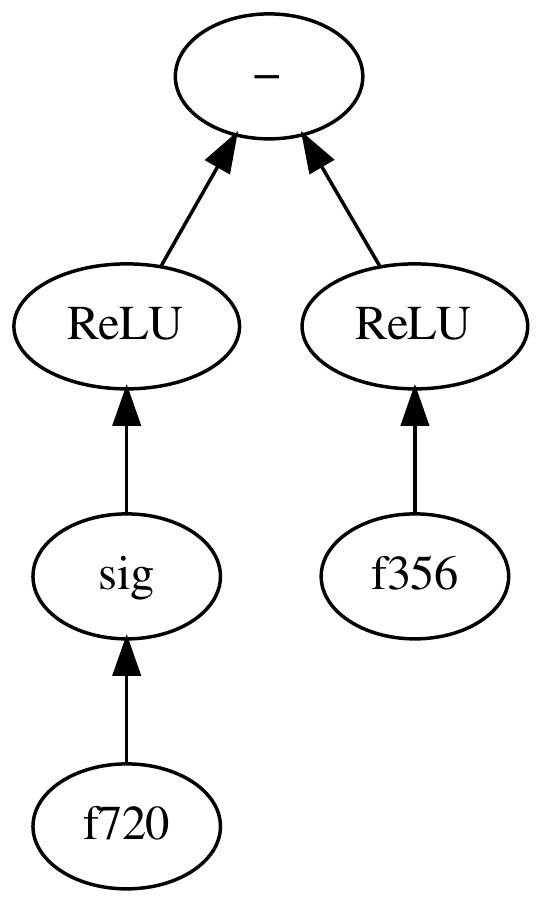}
		\subcaption{}
	\end{subfigure}%
%	\hfill
		\begin{subfigure}[b]{.15\linewidth}
		\includegraphics[width=\linewidth]{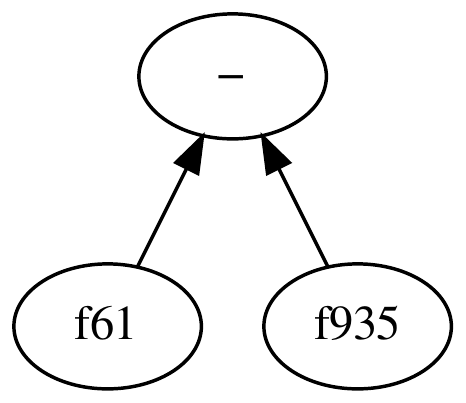}
		\subcaption{}
	\end{subfigure}
%	\hfill
	\begin{subfigure}[b]{.26\linewidth}
		%\vspace{-1em}
	\includegraphics[width=\linewidth]{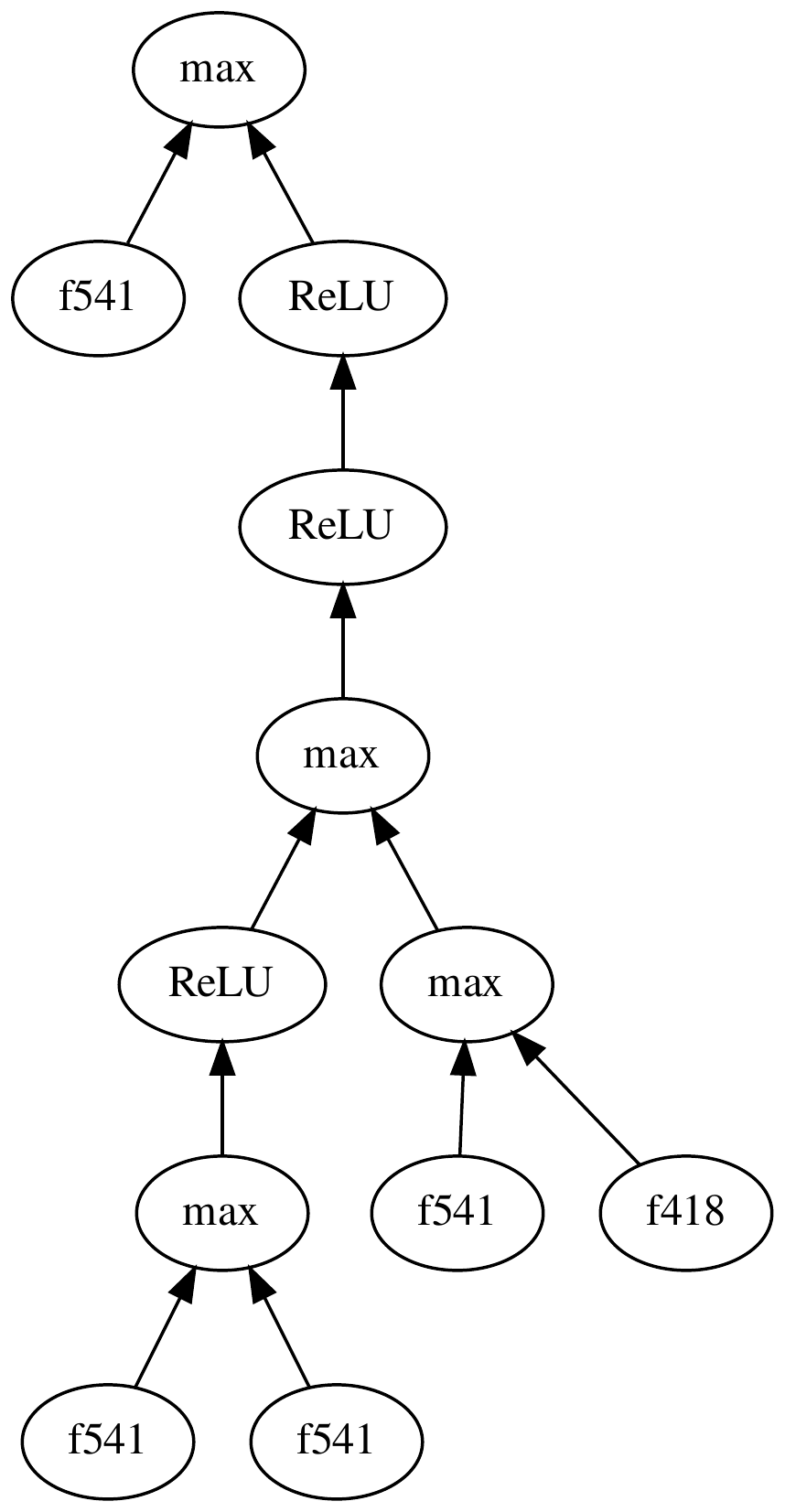}
	\subcaption{}
\end{subfigure}%
%\hfill	
\begin{subfigure}[b]{.46\linewidth}
	\includegraphics[width=\linewidth]{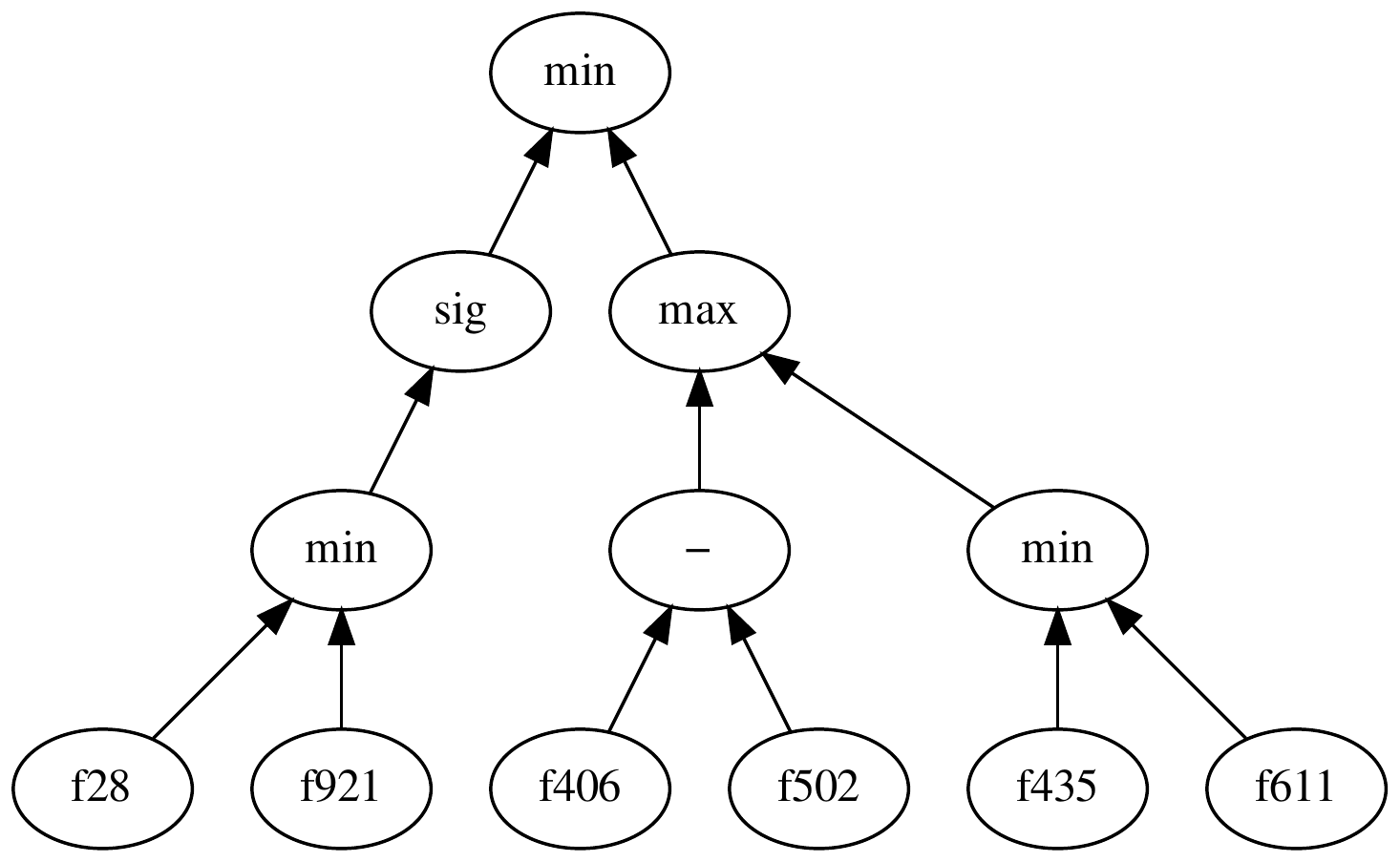}
	\subcaption{}
\end{subfigure}%

\begin{subfigure}[b]{.07\linewidth}
	\includegraphics[width=\linewidth]{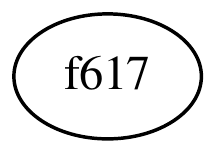}
	\subcaption{}
\end{subfigure}%
\hfill	
\begin{subfigure}[b]{.07\linewidth}
	\includegraphics[width=\linewidth]{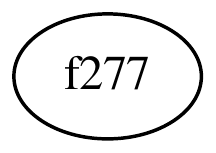}
	\subcaption{}
\end{subfigure}%
\hfill	
\begin{subfigure}[b]{.07\linewidth}
	\includegraphics[width=\linewidth]{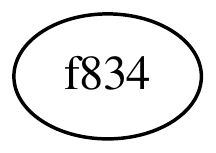}
	\subcaption{}
\end{subfigure}%
\hfill	
\begin{subfigure}[b]{.07\linewidth}
	\includegraphics[width=\linewidth]{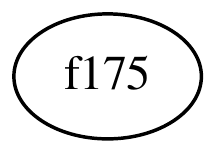}
	\subcaption{}
\end{subfigure}%
\hfill	
\begin{subfigure}[b]{.07\linewidth}
	\includegraphics[width=\linewidth]{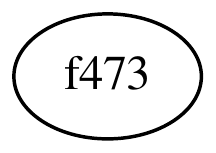}
	\subcaption{}
\end{subfigure}%
\hfill	
\begin{subfigure}[b]{.07\linewidth}
	\includegraphics[width=\linewidth]{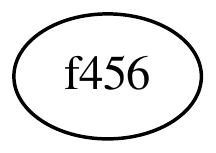}
	\subcaption{}
\end{subfigure}%
\hfill	
\begin{subfigure}[b]{.07\linewidth}
	\includegraphics[width=\linewidth]{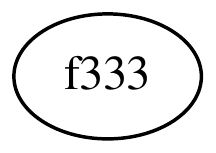}
	\subcaption{}
\end{subfigure}%
\hfill	
\begin{subfigure}[b]{.07\linewidth}
	\includegraphics[width=\linewidth]{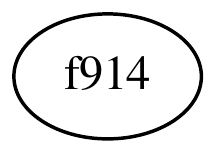}
	\subcaption{}
\end{subfigure}%
\hfill	
\begin{subfigure}[b]{.07\linewidth}
	\includegraphics[width=\linewidth]{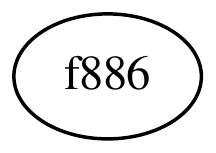}
	\subcaption{}
\end{subfigure}%
\hfill	
\begin{subfigure}[b]{.07\linewidth}
	\includegraphics[width=\linewidth]{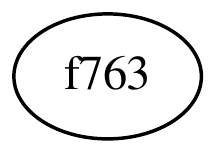}
	\subcaption{}
\end{subfigure}%
\hfill	
\begin{subfigure}[b]{.07\linewidth}
	\includegraphics[width=\linewidth]{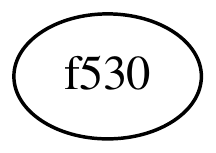}
	\subcaption{}
\end{subfigure}%
\hfill	
\begin{subfigure}[b]{.07\linewidth}
	\includegraphics[width=\linewidth]{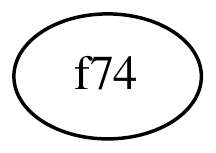}
	\subcaption{}
\end{subfigure}

\begin{subfigure}[b]{.07\linewidth}
	\includegraphics[width=\linewidth]{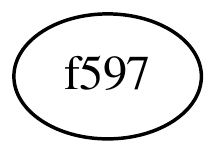}
	\subcaption{}
\end{subfigure}%
\hfill	
\begin{subfigure}[b]{.07\linewidth}
	\includegraphics[width=\linewidth]{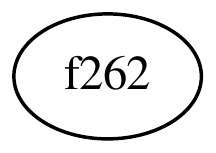}
	\subcaption{}
\end{subfigure}%
\hfill	
\begin{subfigure}[b]{.07\linewidth}
	\includegraphics[width=\linewidth]{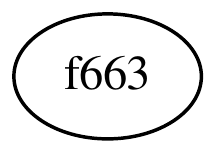}
	\subcaption{}
\end{subfigure}%
\hfill	
\begin{subfigure}[b]{.07\linewidth}
	\includegraphics[width=\linewidth]{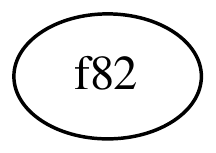}
	\subcaption{}
\end{subfigure}%
\hfill	
\begin{subfigure}[b]{.07\linewidth}
	\includegraphics[width=\linewidth]{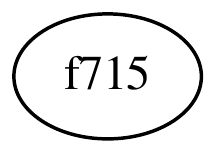}
	\subcaption{}
\end{subfigure}%
\hfill	
\begin{subfigure}[b]{.07\linewidth}
	\includegraphics[width=\linewidth]{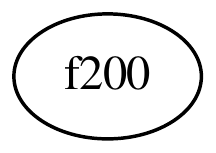}
	\subcaption{}
\end{subfigure}%
\hfill	
\begin{subfigure}[b]{.07\linewidth}
	\includegraphics[width=\linewidth]{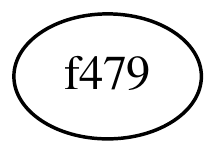}
	\subcaption{}
\end{subfigure}%
\hfill	
\begin{subfigure}[b]{.07\linewidth}
	\includegraphics[width=\linewidth]{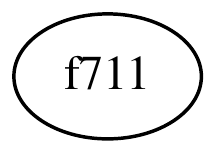}
	\subcaption{}
\end{subfigure}%
\hfill	
\begin{subfigure}[b]{.07\linewidth}
	\includegraphics[width=\linewidth]{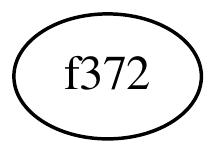}
	\subcaption{}
\end{subfigure}%
\hfill	
\begin{subfigure}[b]{.07\linewidth}
	\includegraphics[width=\linewidth]{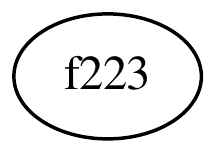}
	\subcaption{}
\end{subfigure}%
\hfill	
\begin{subfigure}[b]{.07\linewidth}
	\includegraphics[width=\linewidth]{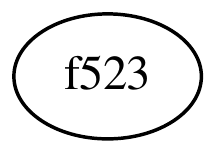}
	\subcaption{}
\end{subfigure}%
\hfill	
\begin{subfigure}[b]{.07\linewidth}
	\hfill
	%\includegraphics[width=\linewidth]{figures/COIL20-28}
	%\subcaption{}
\end{subfigure}%
	\caption{A GP-MaL-MO individual containing 29 evolved trees on the 1024-dimensional COIL20 dataset, with 99.1\% test accuracy using the Random Forest classification algorithm. }\label{example1}
\end{figure}

Of the \revisionOne{11} datasets considered in this paper, the COIL20 dataset has the largest-equal number of features (1,024) and the highest number of classes (at 20). As such, it can be considered to be the ``biggest'' problem: that is, the one that can benefit most greatly from effective manifold learning. An example GP individual trained on the COIL20 dataset, shown in \cref{example1}, has 29 trees (3\% of the original dimensionality) and a very low cost of 0.035. On 10-fold cross validation, it achieves 99.1\% test accuracy using RF, and 96.6\% test accuracy using KNN (higher than when using all features). Even with this impressive accuracy, the evolved trees are surprisingly simple: 23 of them are single selected features, and the other 6 use only 24 \revisionOne{additional} unique features from the original feature set. These 29 trees use only 47 out of 1024 original features (4.6\%), which is clearly a significant reduction in dimensionality. The three most commonly used features are f808 (9 times), f151 (6 times) and f541 (4 times). These three pixels in the original COIL20 images are likely to be particularly characteristic in separating the 20 classes. 

%/home/lensenandr/PycharmProjects/postdoc/src/many_trees_to_plot.py
%/local/scratch/lensenandr/gpemFiles/none/COIL20gpmalmo.gpmal_nc/12/COIL20-0.025562798084805327-29.0.tree
%/local/scratch/lensenandr/gpemFiles/none/mfatCombinedgpmalmo.gpmal_nc/9/mfatCombined-0.0334866714082504-12.0.tree

\begin{figure}
	\renewcommand\thesubfigure{\arabic{subfigure}}
	
	\begin{subfigure}[b]{.22\linewidth}
		\includegraphics[width=\linewidth]{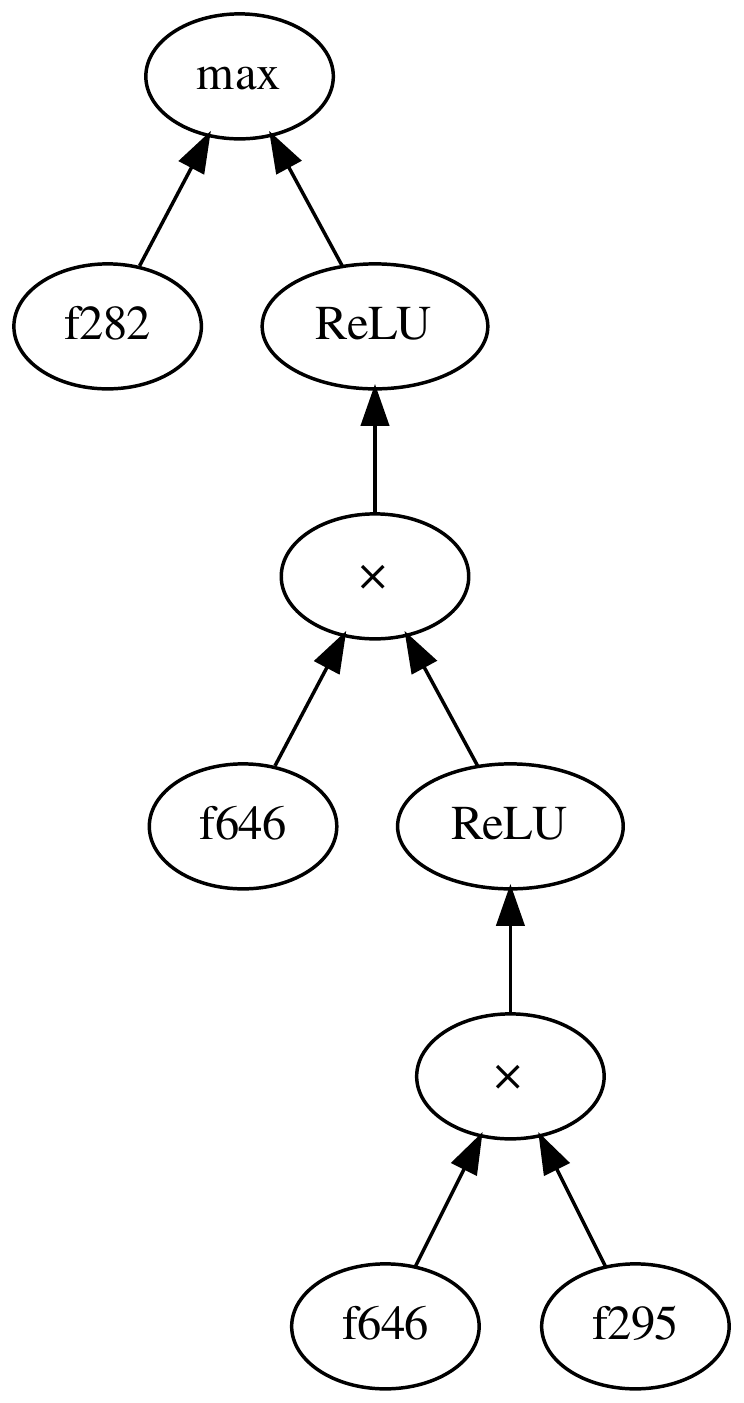}
		\subcaption{}
	\end{subfigure}%
\hfill
	%\begin{subfigure}[b]{.07\linewidth}
		\begin{subfigure}[b]{.07\linewidth}
				\includegraphics[width=\linewidth]{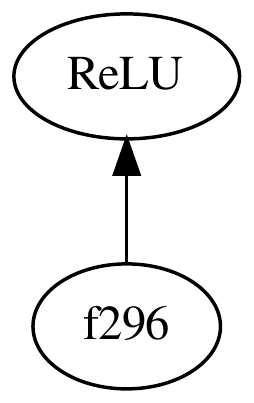}
		\subcaption{}	
	\end{subfigure}
	\hfill
			\begin{subfigure}[b]{.07\linewidth}
		\includegraphics[width=\linewidth]{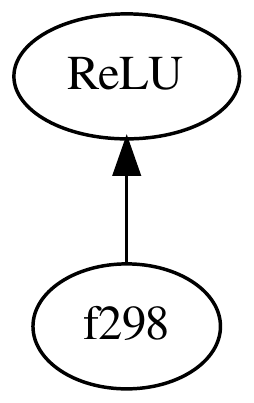}
			\subcaption{}
%	\end{subfigure}
\end{subfigure}
\hfill
	\begin{subfigure}[b]{.07\linewidth}
	\begin{subfigure}{\linewidth}
		\includegraphics[width=\linewidth]{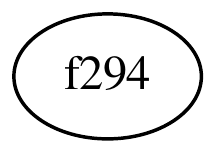}
	\end{subfigure}
	\subcaption{}
	\begin{subfigure}[b]{\linewidth}
		\includegraphics[width=\linewidth]{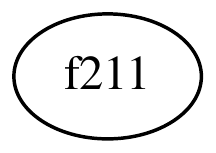}
		\subcaption{}
	\end{subfigure}
	\begin{subfigure}[b]{\linewidth}
	\includegraphics[width=\linewidth]{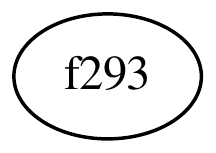}
	\subcaption{}
\end{subfigure}%
\end{subfigure}
\hfill
	\begin{subfigure}[b]{.07\linewidth}
	\begin{subfigure}{\linewidth}
		\includegraphics[width=\linewidth]{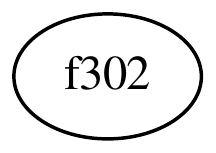}
	\end{subfigure}
	\subcaption{}
	\begin{subfigure}[b]{\linewidth}
		\includegraphics[width=\linewidth]{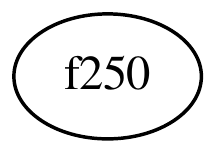}
		\subcaption{}
	\end{subfigure}
	\begin{subfigure}[b]{\linewidth}
		\includegraphics[width=\linewidth]{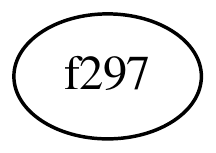}
		\subcaption{}
	\end{subfigure}%
\end{subfigure}%
\hfill
\begin{subfigure}[b]{.07\linewidth}
	\begin{subfigure}{\linewidth}
		\includegraphics[width=\linewidth]{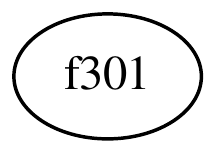}
	\end{subfigure}
	\subcaption{}
	\begin{subfigure}[b]{\linewidth}
		\includegraphics[width=\linewidth]{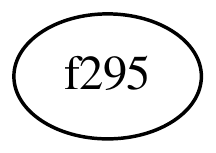}
		\subcaption{}
	\end{subfigure}
	\begin{subfigure}[b]{\linewidth}
		\includegraphics[width=\linewidth]{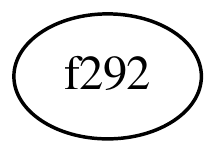}
		\subcaption{}
	\end{subfigure}%
\end{subfigure}%
	\caption{A GP-MaL-MO individual containing 12 evolved trees on the 649-dimensional MFAT dataset, with 94.75\% test accuracy using the Random Forest classification algorithm. }\label{example2}
\end{figure}

On the MFAT dataset, a very simple but powerful individual was found containing 12 trees, as shown in \cref{example2}. The embedding produced by this individual has a cost of 0.033 and achieves 94.75\% test accuracy on RF, and 95.8\% using KNN. Only 13 of the 649 features are used, and the most complex of the trees uses only six functions and three unique features. This model could be used very efficiently on future data (without retraining), as it it requires very little computation to map the 649-dimensional feature space into the 13-dimensional embedding. This model could also be potentially used to give significant insight into the MFAT dataset, by further examining the 13 features used to see how they separate the data into classes. 

Both these two example individuals show the potential of GP-MaL-MO to not only produce representative embeddings at relatively low dimensionality, but also for the evolved mapping to be interpretable and computationally simple. This is a benefit of GP-MaL-MO that is not applicable to any of the other baseline MaL methods. In future work, we plan to investigate this further, by introducing parsimony pressure or a third conflicting objective that minimises the size/complexity of the learned GP trees.

%mfatCombinedgpmalmo.gpmal_nc/29/mfatCombined-0.03508429587482159-11.0.csv 
%COIL20gpmalmo.gpmal_nc/28/COIL20-0.03141250279454488-19.0.tree
%COIL20gpmalmo.gpmal_nc/12/COIL20-0.025562798084805327-29.0.tree
%\newpage
\section{Conclusions}
While manifold learning methods have made a great deal of progress over the last decade, there has been very little focus on mapping-based methods. These methods have a number of appealing traits, such as the ability to be re-used on new data and to provide insight by modelling an embedded manifold based on the original features. In this paper, we significantly extended our previous work --- which was the first use of GP for manifold learning \cite{lensen2019can} --- with a new multi-objective approach that can automatically balance the competing objectives of manifold quality and dimensionality, called GP-MaL-MO. Specialised crossover and mutation operators were designed that allowed for GP individuals to lose and gain trees, as well as mate with individuals of different arity. We also further refined our previous measure of manifold quality. 

Comprehensive testing of GP-MaL-MO showed that it was able to automatically find clear trade-offs between increasing manifold quality and dimensionality. The specialised crossover and mutation operators allowed the proposed method to scale upwards of 100 trees, while the multi-objective algorithm prevented the maximum number of trees being larger than necessary. GP-MaL-MO was shown to be competitive with a range of existing manifold learning methods across a variety of datasets, despite using a more indirect mapping-based approach and using only a subset of the feature set in every tree. Further analysis showed that GP-MaL-MO was able to create interpretable and very efficient models that used a minimal number of features and functions within GP trees. This is a clear advantage of GP-MaL-MO in contrast to other MaL methods.

Future investigation is needed into how best to optimise the Pareto front formed by the two conflicting objectives of manifold quality and dimensionality. While GP-MaL-MO performs well at higher manifold dimensionality, further advancements are needed to improve its performance on small numbers of trees. Promising directions include biasing the MO algorithm towards smaller numbers of trees; utilising some form of adaptive evolutionary pressure; or developing new tailored crossover and mutation operators that pressure solutions towards using a smaller number of more powerful trees. Another direction that we hope to pursue in the future is to automatically encourage simpler, more efficient trees, perhaps by introducing a third competing objective to the EMO algorithm.

Beyond this, there are a number of other open questions in evolutionary manifold learning. The use of GP-based MaL to perform visualisation (i.e.\ two-dimensional manifold learning) could be promising if it could produce interpretable trees that give accurate and understandable visualisations. Embedding MaL methods traditionally use gradient descent to optimise the embedded space. Given that this is a form of numerical optimisation, it seems likely that an EC-based numerical optimisation method such as Differential Evolution (DE) \cite{price2013de} could give superior results that are less prone to local optima. 
%\begin{itemize}
%	\item First MO approach
%	\item Competitive with existing methods?
%	\item More work needed to improve performance at small numbers of trees.
%\end{itemize}

%\begin{acknowledgements}
%\end{acknowledgements}

\section*{Conflict of Interest}
The authors declare that they have no conflict of interest.

\bibliographystyle{spmpsci}      % mathematics and physical sciences
\bibliography{biblo}

\end{document}